\documentclass[runningheads]{llncs}

\usepackage[mobile]{eccv}

\usepackage{eccvabbrv}

\usepackage{graphicx}
\usepackage{booktabs}

\usepackage[accsupp]{axessibility}  

\usepackage{hyperref}

\usepackage{orcidlink}

\usepackage{multicol}
\usepackage{multirow}
\usepackage{pifont}
\newcommand{\squishlist}{
 \begin{list}{$\bullet$}
  { \setlength{\itemsep}{0pt}
     \setlength{\parsep}{1pt}
     \setlength{\topsep}{1pt}
     \setlength{\partopsep}{0pt}
     \setlength{\leftmargin}{1.5em}
     \setlength{\labelwidth}{1em}
     \setlength{\labelsep}{0.5em} } }
\newcommand{\squishend}{
  \end{list}  }
\newcommand{\red}[1]{\textcolor{red}{#1}}
\newcommand{\blue}[1]{\textcolor{blue}{#1}}
\newcommand{\gray}[1]{\textcolor{gray}{#1}}

\usepackage[misc]{ifsym}
\newcommand{\envelope}{\ding{41}}

\usepackage{wrapfig}
\DeclareMathOperator*{\softmax}{softmax}

\newcommand{\rowname}[1]
{\rotatebox{90}{\makebox[1.0\tempheight][t]{~~\quad{\textbf{#1}}}}}
\begin{document}

\title{Learning to Detect Multi-class Anomalies with Just One Normal Image Prompt} 

\titlerunning{OneNIP}

\author{\textbf{Bin-Bin Gao}\orcidlink{0000-0003-2572-8156} \inst{\text{\envelope}}}

\authorrunning{B.-B. Gao}

\institute{
Tencent YouTu Lab \\
\email{csgaobb@gmail.com} \\
Code and Models: \url{https://github.com/gaobb/OneNIP}
}

\maketitle
\let\thefootnote\relax\footnotetext{$^\text{\envelope}$ B.-B. Gao is the corresponding author.}

\begin{abstract}
  Unsupervised reconstruction networks using self-attention transformers have achieved state-of-the-art performance for multi-class (unified) anomaly detection with a single model. However, these self-attention reconstruction models primarily operate on target features, which may result in perfect reconstruction for both normal and anomaly features due to high consistency with context, leading to failure in detecting anomalies. Additionally, these models often produce inaccurate anomaly segmentation due to performing reconstruction in low spatial resolution latent space. To enable reconstruction models enjoying high efficiency while enhancing their generalization for unified anomaly detection, we propose a simple yet effective method that reconstructs normal features and restores anomaly features with just One Normal Image Prompt (OneNIP). In contrast to previous work, OneNIP allows for the first time to reconstruct or restore anomalies with just one normal image prompt, effectively boosting unified anomaly detection performance. Furthermore, we propose a supervised refiner that regresses reconstruction errors by using both real normal and synthesized anomalous images, which significantly improves pixel-level anomaly segmentation. OneNIP outperforms previous methods on three industry anomaly detection benchmarks: MVTec, BTAD, and VisA. 
  
\end{abstract}    
\section{Introduction}
\label{sec:intro}

Unsupervised visual anomaly detection aims to learn models only on normal training samples and expects these learned models to be capable of detecting anomalies at the image level and even localizing anomaly regions at the pixel level for both normal and anomaly testing samples. 
Anomaly detection (AD) has a wide range of applications, including video surveillance~\cite{sultani2018real,learningmemory,georgescu2021anomaly}, medical image diagnosis~\cite{kim2022feasibility,xiang2021painting}, industrial defect inspection in manufacturing~\cite{mvtec,btad,bergmann2022beyond}, and more. Most AD methods~\cite{patchcore,csflow,padim,dfm,cfa,draem,sspcab,rd,simplenet} mainly focus on training separated models for different objects or textures. 
However, this separated paradigm (one model for one class) may be not practical, as it requires high memory consumption and storage burden, especially when the number of classes increases. 
In contrast, unified AD (one model for all classes) attempts to detect various anomalies for all categories using a single model. 
Compared to the separated training mode, the unified AD paradigm is more challenging as it requires handling more complex data distributions. Therefore, most AD methods often suffer from a significant performance drop when extending them from separated paradigms to unified ones. Furthermore, it seems necessary to study unified AD from a foundational model perspective.

\begin{figure}[tb]
  \centering
  \begin{subfigure}{0.45\linewidth}
  \includegraphics[width=1.00\columnwidth,keepaspectratio]{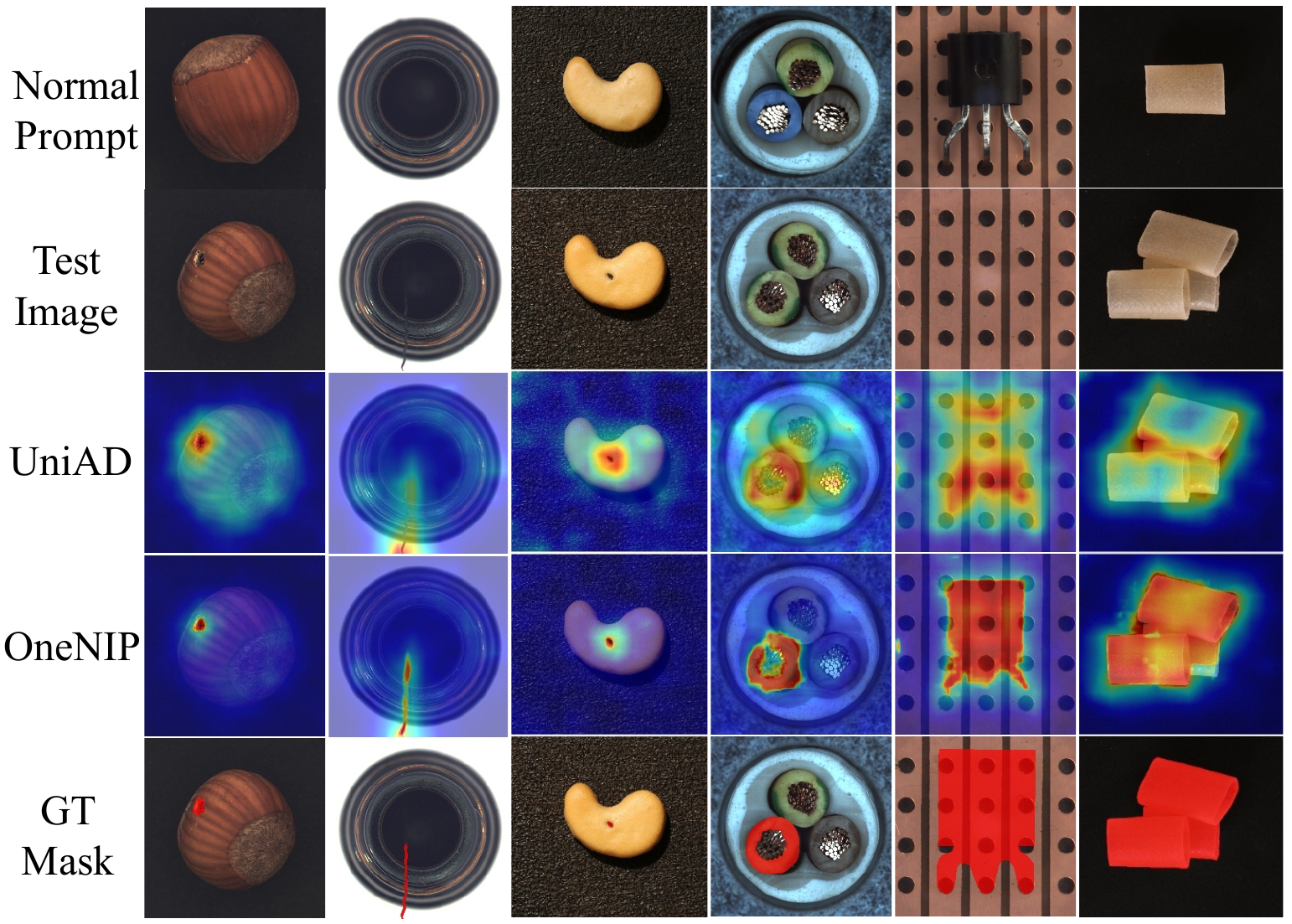} 
    \caption{Qualitative comparisons on selected common (left three columns) and camouflaged (right three columns) anomaly images.}
    \label{fig:uniadvsonenip-a}
  \end{subfigure}
  \hfill
  \begin{subfigure}{0.52\linewidth}
    \includegraphics[width= 1.00\columnwidth]{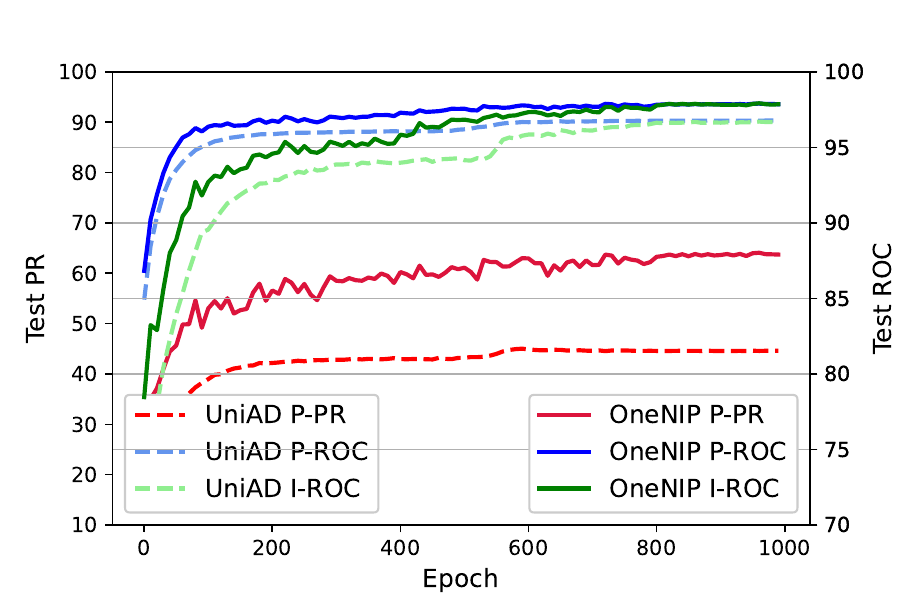} 
    \caption{Testing metrics (I-ROC, P-ROC and P-PR) comparisons as a function of training epoch on MVTec testing set.}
    \label{fig:uniadvsonenip-b}
  \end{subfigure}
  \vspace{-5pt}
  \caption{\textbf{Comparisons of state-of-the-art UinAD and our OneNIP.} The proposed OneNIP detects anomalies through learning comparison with one normal image as a visual prompt. Compared to UniAD, OneNIP enjoys more accurate anomaly localization (a) and faster convergence (b).}
  \label{fig:uniadvsonenip}
  \vspace{-5pt}
\end{figure}

A recent work (UniAD)~\cite{uniad} attempts to detect multiple anomalies for all categories with a unified model using a transformer reconstruction network. However, the pure transformer suffers from over-fitting because of ``identity shortcut'' issue, which appears as returning a direct copy of input disregarding its content. This implies that even anomalous samples can be well recovered with the learned model and hence fail to be detected. To address this issue, UniAD~\cite{uniad} proposed a layer-wised query decoder and a neighbor-masked attention (NMA) to prevent model learning from the shortcut. Similar to NMA in UniAD, SSPCAB~\cite{sspcab} learns to reconstruct the masked area using contextual information implemented by dilated convolutional. Despite UniAD and SSPCAB employing different architectures and implementation strategies, they share the same spirit of reconstruction with context. In this way, the performance of AD can be ensured as most objects inherently possess specific physical structures or geometric characteristics as shown in \cref{fig:uniadvsonenip-a} (left three columns). However, for some complex scenarios, e.g., camouflaged anomalies (right three columns in \cref{fig:uniadvsonenip-b}), which refer to abnormal regions that are “seamlessly” embedded in their context in an image, it is hard to effectively detect them only using contextual information of themselves.

In order to explore a more general anomaly detection, let's first recall how we humans recognize anomalies. 
Generally, people are able to perceive anomalies when an input significantly deviates from those normal or expected patterns stored in the human brain. There actually, in fact, exists evidence to support this point in neural science. For example, predictive coding theory states that the human brain compares its expectations with the data it receives, and sends discrepancies (prediction errors) to higher levels~\cite{predictivecoding}. 
This process allows the brain to perceive anomalies based on memory and contextual information. PatchCore~\cite{patchcore} indeed captures normal local patch features, stores them in a memory bank, and then recognizes anomalies by comparing input features with the memory bank. In addition, some distribution-based methods~\cite{padim,dfm} model a multivariate Gaussian distribution for normal local features, then utilize a distance metric to measure anomalies. However, these memory- and distribution-based methods still struggle with detecting camouflaged anomalies because of ignoring global structuration information.

Naturally, we raise a question: how to elegantly leverage both contextual and global structural information to enhance the performance of anomaly detection? In this paper, we propose a simple yet effective anomaly detection framework that utilizes a normal image as a global prompt to guide the feature reconstruction, which is inspired by predictive coding theory~\cite{predictivecoding}. Under the guidance of a normal image prompt, a feature reconstruction network can leverage self-attention mechanisms to model contextual information, while also conveniently facilitating interaction between target feature and global image prompt using cross-attention. Therefore, our approach can effectively detect both common and camouflaged anomalies by utilizing a normal image prompt as shown in \cref{fig:uniadvsonenip-a}. Compared to state-of-the-art UniAD, our method exhibits faster convergence as shown in \cref{fig:uniadvsonenip-b}.
Our contributions are summarized as follows:

\squishlist 
\item We propose a novel unified anomaly detection framework, that unsupervised reconstructs normal features utilizing both contextual information themselves and corresponding global information from a normal image prompt.

\item To enhance the guidance of the normal image prompt, we introduce pseudo-anomalous samples and propose an unsupervised restoration stream that pushes these pseudo features to recover to their corresponding normal ones.

\item We propose a supervised refiner that regresses reconstruction errors from low to high resolution with both real normal samples and pseudo-anomalous samples, which greatly boosts the performance of anomaly segmentation.

\item Our method achieves state-of-the-art performance with a unified setting on three industry anomaly detection benchmarks, MVTec, BTAD, and VisA.
\squishend

\section{Related Work}\label{sec:related}

\textbf{Embedding-based AD methods} leverage offline features extracted from pre-trained models for anomaly detection. It assumes that these offline features preserve discriminative information and thus help to separate anomalies from normal samples. PaDiM~\cite{padim}, MDND~\cite{modelingdistribution}, and DFM~\cite{dfm} model a normal distribution based on normal features, then utilize a distance metric to measure anomalies. PatchCore~\cite{patchcore} captures normal features and stores them in a memory bank, and calculates anomaly scores by comparing all patch features and the memory bank. However, computing the inverse of covariance in the normal distribution or searching in the memory bank brings large memory-consuming. In addition, there is a domain gap between target (industrial images) and source distribution (e.g., ImageNet) if directly using offline features. 
CS-Flow~\cite{csflow} proposes to transform normal feature distribution into Gaussian distribution via normalizing flow. Further, PyramidFlow~\cite{pyramidflow} combines latent templates and normalizing flow for high-resolution anomaly localization. CFA~\cite{cfa} and PADA~\cite{panda} propose feature adaptation for adapting targeted datasets. Knowledge distillation methods~\cite{us,glancing,stfpm,us,mkd,rd,rd++} train a student network to match a fixed pre-trained teacher network. However, they always are limited by designing structural differences between teacher and student. 

\noindent\textbf{Discriminator-based methods} typically convert unsupervised anomaly detection to supervised anomaly detection by introducing pseudo~(synthesized) anomaly samples. CutPaste~\cite{cutpaste} proposes a simple strategy to generate synthetic anomalies, which cuts a small rectangular area of variable sizes and aspect ratios from normal training images and pastes this patch back to the image at a random location. Similar to CutPaste, DRAEM~\cite{draem} generates pseudo anomaly images using Perlin~\cite{perlin} and obtains binarized anomaly maps. CutPaste~\cite{cutpaste} learns an image-level classifier for enhancing discrimination between normal and anomaly features, while DRAEM~\cite{draem} learns an additional pixel-level segmentation model with pseudo-mask. PRN~\cite{prn} presents a variety of anomaly generation strategies for more accurate anomaly localization. DeSTSeg~\cite{destseg} proposes a denoising knowledge distillation and employs a segmentation network for accurate anomaly localization with synthetic samples. BGAD~\cite{bgad} proposes a boundary-guided semi-push-pull loss
for learning more discriminative features with normal and synthetic samples. Instead of synthesizing anomalies on images, SimpleNet~\cite{simplenet} generates anomaly features by adding Gaussian noise to normal features and then learns a binary discriminator to distinguish anomaly features from normal ones. \cite{chiu2023self} proposes a self-supervised normalizing flow-based density estimation model, which is
trained by normal images and synthetic anomalous images.

\noindent\textbf{Reconstruction-based AD methods} assume that anomalous image regions or features should not be able to be properly reconstructed since they do not exist in normal training samples. Some works use generative models such as auto-encoders~\cite{demae,gdae,memorize,l2ae_ssimae,divideassemble,lee2022anovit} and generative adversarial networks~\cite{ocgan,scadn,oldgold} to reconstruct normal images. RGI~\cite{rgi} proposes a robust GAN-inversion that can restore any input image~(even with gross corruptions) to a clean image and identify the corrupted region mask by solving the optimization problems thereof. Some works frame anomaly detection as an inpainting problem, where patches from images are partly masked. RIAD~\cite{zavrtanik2021reconstruction} randomly removes partial image regions and reconstructs the image from partial inpaintings with a convolutional neural network. SSPCAB~\cite{sspcab} learns to reconstruct masked regions using contextual information with a masked convolutional kernel. To enhance reconstruction diversity while avoiding the undesired generalization of anomalies, a pyramid deformation module is proposed to model diverse normal and measure the severity of anomaly in \cite{diversitymeas}. These methods tend to be computationally expensive because they involve reconstruction in image space. The recent UniAD~\cite{uniad}, omniNAL~\cite{omnial} and FOD~\cite{yao2023focus} reconstruct features extracted from a pre-trained model and achieve state-of-the-art performance for unified anomaly detection. However, pixel-level anomaly segmentation is still unsatisfactory.

\noindent\textbf{Prompt-based AD methods} using large pre-trained vision-language models, e.g., CLIP~\cite{clip}, have shown unprecedented generality, and achieve impressive performance on various tasks, such as zero- and few-shot image classification, open-vocabulary object detection~\cite{du2022learning}, and text-to-image generation~\cite{ldm}. Recent studies, WinCLIP~\cite{winclip}, SAA+~\cite{cao_segment_2023}, AnomalyCLIP~\cite{zhou2023anomalyclip} and MVFA~\cite{mvfa2024}, have demonstrated that utilizing multiple fixed textual prompts or learning a dynamic textual prompt on a powerful CLIP model~\cite{clip} can yield excellent performance for zero- and few-shot anomaly detection. 
Furthermore, AnomalyGPT~\cite{anomalygpt} applying multi-turn dialogues not only indicates the presence and location of the anomaly but also provides a detailed description of the anomaly in a testing image. However, these methods primarily rely on textual prompts to identify anomalies. Different from them, we explore to detect anomalies via a normal image as a visual prompt.

\section{Methods}\label{sec:method}

\begin{figure*}[t]
    \centering
    \includegraphics[width=0.96\linewidth]{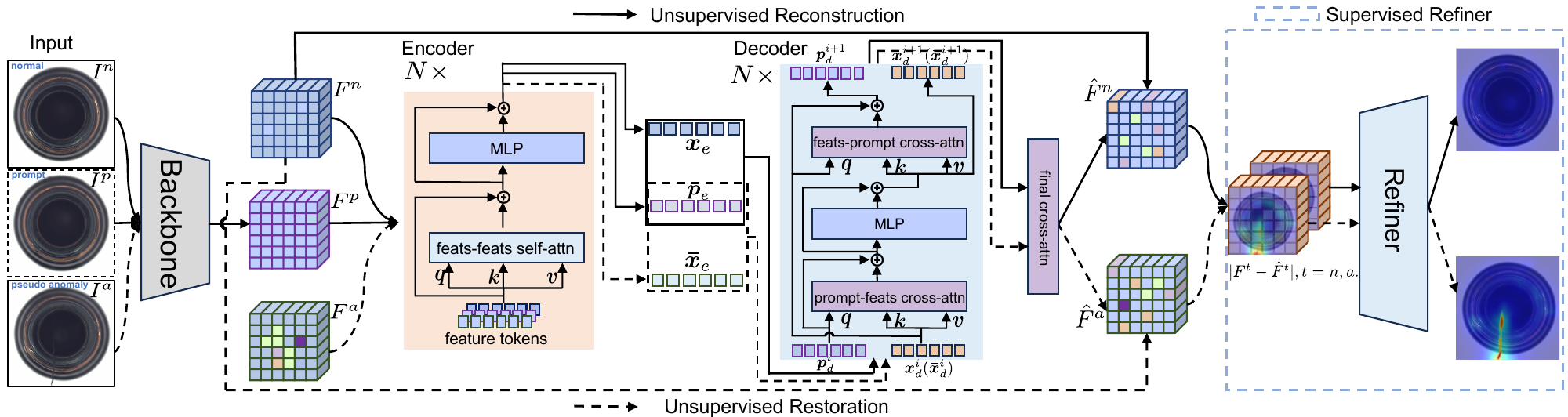}
    \caption{\small{\textbf{Overview of OneNIP for unified anomaly detection.} In the training stage, both normal and synthetic images are fed a pre-trained backbone for extracting multi-level representation. Under the guidance of a normal image prompt, the normal features are reconstructed in an unsupervised reconstruction stream~(\cref{sec:rec}), and the synthetic anomaly features are restored in an unsupervised restoration stream~(\cref{sec:res}). Furthermore, a supervised refiner~(\cref{sec:ref}) is used to regress reconstruction errors for both normal and synthetic anomaly images. The unsupervised restoration stream will be removed at inference.}}
    \label{fig:framework}
    \vspace{-5pt}
\end{figure*}

Our OneNIP is built on state-of-the-art UniAD and is mainly composed of an unsupervised reconstruction, an unsupervised restoration, and a supervised refiner as shown in Fig.~\ref{fig:framework}. The unsupervised reconstruction and unsupervised restoration share the same encoder-decoder architecture. The encoder models contextual information with a self-attention transformer, while the decoder models the relationship between target features and a normal prompt with a bidirectional cross-attention transformer. The supervised refiner regresses the reconstruction errors from low to high resolution for more accurate anomaly localization.

Concretely, for a normal input image $I^n$ and its corresponding prompt image $I^p$ ($I^n$ and $I^p \in \mathcal R^{H\times W\times3}$), we parallelly extract their offline features ($F^n$ and $F^p \in \mathcal R^{h\times w\times c} $) using a pre-trained backbone, e.g., EfficientNet-b4~\cite{efficientnet}. Then, the self-attention encoder independently processes them with added positional embeddings for modeling contextual dependencies. Next, these encoded features and prompt tokens are dynamically updated in two directions (prompt-to-features and features-to-prompt) with a bidirectional decoder consisting of multiple two-way cross-attention blocks. We expect the original feature $F^n$ to be well reconstructed by fully exploring both the context and relationship of the target feature and normal image prompt. To further enhance the guidance of the normal image prompt, we introduce pseudo-anomalous image $I^a$ synthesizing from $I^n$ and propose an unsupervised feature restoration stream that pushes the pseudo-anomalous feature $F^a$ to recover to its corresponding normal feature by refusing the encoder-decoder network. At inference, the unsupervised restoration
stream can be flexibly removed. Finally, the reconstruction errors between the original and reconstructed ones are refined by a lightweight supervised refiner module with real normal and pseudo anomaly samples and their pixel-level anomaly masks. Next, we elaborately introduce them in this section. 

\subsection{Reconstruction with Normal Image Prompt}\label{sec:rec}

\noindent\textbf{Revisiting UniAD.}
We focus on unified anomaly detection which requires a model to handle more complex data distribution and thus is more challenging. As we know, reconstruction-based UniAD~\cite{uniad} using an encoder-decoder transformer is a powerful and state-of-the-art solution for unified anomaly detection and is composed of neighbor-masked attention (NMA) and layer-wise query decoder (LQD). The NMA limits that one feature can't see itself and its neighborhoods and thus takes its contextual features (long dependencies) for reconstruction. On the other hand, a learnable query embedding $\vec {q}^i$ is first fused with the encoder embedding $\vec {x}_e$ and then integrated with the outputs $\vec {x}^i_d$ of the $i$-th decoder layer. Experiments have proven that the learned query embedding can alleviate over-fitting. 
For simplicity, we here omit MLP, residual connection, layer normalization, and dropout in LQD, and then formulate two important steps in the $i$-th block of LQD as follows: 
\begin{equation}\label{eq:lqd}
\begin{split}
\vec {q}\prime  = &\softmax(\vec q^i {\vec {x}_e}^T/{\sqrt{c}}) \vec {x}_e, \\
\vec {x}^{i+1}_d  = &\softmax(\vec {q}\prime {\vec {x}^i_d}^T/{\sqrt{c}}) \vec {x}^i_d, 
\end{split}
\end{equation}
where $c$ is the dimension of $\vec {x}_e$  and $\vec {x}^0_d$ is initialized by $\vec x_e$ at the first block in LQD. In actual experiments, Eq.~\ref{eq:lqd} is implemented by multi-head self-attention~\cite{attention}. In each block of LQD, it is important to note that $\vec q^i$ is individual and learnable, while $\vec {x}_e$ remains fixed and unchanged. The LQD reconstructs features only by themselves, which may lead to failure when facing challenging anomalies.

\noindent\textbf{Unidirectional decoder with static prompt.}
We expect the feature reconstruction not only to rely on its structure and characteristics but also to be guided by a normal prompt, aiming to reduce the difficulty of reconstruction and improve the performance of anomaly detection. A simple and naive manner is to directly replace the query embedding $\vec {q}^i$ in the LQD with the encoder output $\vec {p}_e$ of a normal image prompt $I^p$, thereby enabling the interaction between the prompt and target features. Therefore, we convert Eq.~\ref{eq:lqd} as follows: 
\begin{equation}\label{eq:unid}
\begin{split}
\vec {q}\prime  = &\softmax(\red{\vec {p}_e} {\vec {x}_e}^T/{\sqrt{c}}) \vec {x}_e, \\
\vec {x}^{i+1}_d = &\softmax(\vec {q}\prime {\vec {x}^i_d}^T/{\sqrt{c}}) \vec {x}^i_d.  
\end{split}
\end{equation}
The change is simple but has completely different implications and boosts the performance of anomaly detection (Tab.~\ref{tab:as}). In Eq.~\ref{eq:unid}, the prompt encoding statically interacts with the target feature in a unidirectional manner, hence we call it unidirectional decoder. However, this unidirectional mode may not be flexible enough and may fail to align with the target feature especially when the target feature is continuously updated.

\noindent\textbf{Bidirectional decoder with dynamic prompt.}
Unlike the static prompt in the unidirectional decoder, we dynamically update both prompt and target features using a pair of bidirectional cross-attention as follows:
\begin{equation}\label{eq:bid}
\begin{split}
\red{\vec {p}^{i+1}_d}  = &\softmax(\red{\vec {p}^i_d} \red{\vec {x}^i_d}^T/{\sqrt{c}}) \red{\vec {x}^i_d}, \\
\vec {x}^{i+1}_d  = &\softmax(\vec {x}^i_d {\red{\vec {p}^{i+1}_d}}^T/{\sqrt{c}}) \red{\vec {p}^{i+1}_d},  
\end{split}
\end{equation}
where $\vec {p}^0_d$ and $\vec {x}^0_d$ is initialized by $\vec {p}_e$ and $\vec {x}_e$ from the encoder module.
The above bidirectional decoder models two-directional feature interactions including prompt-to-features and features-to-prompt. The first interaction performs a cross-attention from prompt tokens (as queries) to the target features, and the second interaction performs another cross-attention from the target features (as queries) to prompt tokens.
The next decoder block takes updated prompt tokens and target features from the previous block. In this way, the target feature reconstruction not only utilizes its contextual information but also leverages the corresponding normal prompt dynamically. It is worth noting that this bidirectional modeling way also enhances the flexibility of the prompt features and can adapt to the distribution shift of target features to some extent. Last, a final cross-attention is used to update prompt tokens on the outputs of the bidirectional decoder, and its output is taken as the reconstructed one ($\hat F^n$) of the original feature ($F^n$).
The reconstruction loss function computes the mean squared error (MSE) between the reconstructed and original features as
\begin{equation}\label{eq:unrec}
    \mathcal L_{rec} = \frac{1}{c \times h \times w} \sum_{i=1}^c
    \sum_{j=1}^{h} \sum_{k=1}^{w} (F^n_{i, j, k} - \hat F^n_{i, j, k} )^2.
\end{equation}
\subsection{Restoration with Normal Image Prompt} \label{sec:res}

It can be observed that the unsupervised reconstruction learning is solely performed on normal training images, which may lead the model to rely more on its contextual information and weaken the involvement of the image prompt in the reconstruction process. To further enhance the guidance of the image prompt, an expected manner is to increase the difficulty of the reconstruction task, forcing the network to rely not only on contextual information from itself but also on prompt information from the normal image prompt. To achieve this, we introduce artificially synthesized pseudo anomaly image $I^a$, which can be easily generated by adding corruptions or disruptions to a normal training image $I^n$, such as CutPaste~\cite{cutpaste} and DRAEM~\cite{draem}. Based on the pseudo anomaly image, we can convert the previous reconstruction into a restoration problem that expects to restore the anomaly feature $F^a$ to the normal one $F^n$ with a normal image prompt $I^p$. This restoration manner is consistent with the expectation of reconstruction models during the testing phases.

Similar to the reconstruction process, we first feed a pair of images ($I^a$ and $I^p$) into a pre-trained backbone for extracting offline features ($F^a$ and $F^p$) and then obtain the ultimately repaired features $\hat F^a$ sequentially applying the offline paired features into a self-attention encoder and a bidirectional cross-attention decoder. Specifically,
the self-attention encoder parallelly takes $F^a$ and $F^p$ as inputs and outputs as $\vec {\bar x}_e$ and $\vec {p}_e$. Next, the bidirectional decoder is initialized by $\vec {\bar x}_e$ and $\vec {p}_e$, and dynamically updated with Eq.~\ref{eq:bid}. In the $i$-th block of the bidirectional decoder, we denote dynamic feature and prompt as $\vec {\bar x}^i_d$ and $\vec {p}^i_d$, which is easily obtained by simply replacing $\vec {x}^i_d$ with $\vec {\bar x}^i_d$ in Eq.~\ref{eq:bid}. Different from the objective function Eq.~\ref{eq:unrec} in unsupervised reconstruction, the restoration loss function computes the MSE between the restored feature ($\hat F^a$) and the corresponding original normal feature ($F^n$) as
\begin{equation}\label{eq:unres}
    \mathcal L_{res} = \frac{1}{c \times h \times w} \sum_{i=1}^c
    \sum_{j=1}^{h} \sum_{k=1}^{w} (F^n_{i, j, k} - \hat F^a_{i, j, k} )^2.
\end{equation}

\subsection{Supervised Refiner} \label{sec:ref}
Given a normal training image $I^n$ and the corresponding anomaly mask $M^n$ (all elements are zero), we can synthesize an anomaly image $I^a$ and denote its anomaly mask as $M^a$. Then, we feed the normal and synthesized images $\{I^t\}_{t=\{n,a\}}$ into a pre-trained backbone and derive their offline representation as $\{F^t\}_{t=\{n,a\}}$. Then, we reconstruct the normal $F^n$ as $\hat F^n$ with the proposed reconstruction stream and restore anomaly $F^a$ as $\hat F^a$ with the proposed restoration stream, respectively. Here, we use the absolute element-wise subtraction of original and reconstructed (restored) features to measure their difference, that is
\begin{equation}\label{eq:absdiff}
    E^t = |F^t - \hat F^t|, t= \{n, a\}.
\end{equation}
In fact, the $L_2$ norm of $E^t$ in Eq.~\ref{eq:absdiff} can be used to roughly localize anomaly regions. In this way, however,  it is hard to accurately locate abnormal regions since feature reconstruction or restoration is performed in a low-resolution (i.e., 1/16 of original input) latent space. 

Note that the synthesized anomaly image $I^a$ naturally carries pixel-level anomaly mask $M^a$, we expect to fully utilize $M^a$ to further refine reconstruction errors from low to high resolution. To end this, we design a lightweight and pixel-level refiner based on reconstruction errors for performing anomaly segmentation. The refiner consists of several transposed convolution blocks following a 1$\times$1 convolution layer. Here, each transposed convolution block upsamples the reconstruction error $E^t$ by 2$\times$, and it is composed of a 3$\times$3 convolution, a BatchNorm, a ReLU, and a 2$\times$2 deconvolution. In our experiment, we employ two transposed convolutional blocks and thus upscale the reconstruction error from 1/16 to 1/4 relative to the input image. Finally, the 1$\times$1 convolution layer transforms the channel number of upscaled reconstruction error to 1 and obtains an estimated anomaly map as $\hat M^t$. For compute loss between $\hat M^t$ and the ground-truth $M^t$, we further resize $\hat M^t$ to the size of $M^t$. Considering that anomaly pixels are typically in the minority in anomaly detection, we utilize Dice loss~\cite{wei2021learn}, which is effective for learning from extremely imbalanced data, that is 
\begin{equation}
\mathcal L_{seg} = 1 - \frac{2 \cdot \sum_{i=1}^{H} \sum_{j=1}^{W} \hat M_{i,j}^t \cdot M_{i,j}^t} {\sum_{i=1}^{H} \sum_{j=1}^{W} (\hat M^t_{i,j})^2 + \sum_{i=1}^{H} \sum_{j=1}^{W} (M^t_{i,j})^2 },
\end{equation}
where $(i,j)$ represents a spatial location in $M^t$ or $\hat M^t$.

\subsection{Training Loss}

During training, given an image of a specific class, we randomly sample a normal image among all training images of this class to serve as its prompt by default. 
In addition, we also explore other prompt strategies, e.g., fixed mode,
which means that only one fixed image prompt is used for
each category during training.
Considering all three objectives including unsupervised reconstruction, unsupervised restoration, and supervised refiner in our OneNIP, the total training loss is
\begin{equation}
    \mathcal L = L_{rec} +  L_{res} + \lambda L_{seg},
\end{equation}
where $\lambda > 0$ is a weight that balances the importance of the two types of loss functions $L_{rec} + L_{res}$ and $L_{seg}$.

\subsection{Inference}
At inference, we first randomly select a normal training image for each class and then pre-extract offline prompt features for constructing a class-aware prompt pool $\{P_i\}_{i=1}^{~C}$.  
Given a testing image $I$ and its feature $F$, we can derive an appropriate prompt by computing the cosine similarity between the testing feature $F$ and the prompt pool because the class of the testing image is agnostic.

\noindent\textbf{Pixel-Level Anomaly Segmentation:} 
For unsupervised reconstruction, the anomaly score map is calculated as the $L_2$ norm of the reconstruction error as
\begin{equation}
    S_{rec} = ||F - \hat F||_2 \in \mathbb{R}^{h \times w}.
\end{equation}
For supervised refiner, the anomaly score map is predicted as $\hat M \in \mathbb{R}^{H \times W}$. Finally, we combine $S_{rec}$ (resizing into original resolution) and $\hat M$ together and take it as the final anomaly segmentation map, that is
\begin{equation}
    S = (1-\alpha) \cdot S_{rec} + \alpha \cdot \hat M,
\end{equation}
where $\alpha \in [0,1]$ is a weight.

\noindent\textbf{Image-Level Anomaly Classification:} Anomaly classification aims to detect whether an image contains anomalous regions. Following the previous work, we take the maximum value of $S$ as the image-level anomaly score.

\section{Experiment}\label{sec:exp}

\subsection{Experimental Setup}
Following the previous works, we comprehensively evaluate our method on three industry anomaly detection benchmarks, including {MVTec}~\cite{mvtec}, {BTAD}~\cite{btad}, and VisA~\cite{visa}.

\noindent\textbf{Protocol}: We train a single model for detecting all categories following  UniAD. For fair comparisons, we use the original training/testing splits given in previous works~\cite{mvtec,btad,visa}. In our experiments, all images are resized to 224$\times$224 for training and testing unless otherwise specified.

\noindent\textbf{Metric}: We compare the state-of-the-art anomaly detection methods with our OneNIP using ROC and PR metrics in image- and pixel levels. We argue that the PR metric is better for anomaly segmentation, where the imbalance issue is very extreme between normal and anomaly pixels~\cite{visa,rocpr}.

\noindent\textbf{Comparison Methods}: 
We compare our method with diverse state-of-the-art anomaly detection methods including CS-Flow~\cite{csflow}, PaDiM~\cite{padim}, DFM~\cite{padim}, PatchCore~\cite{patchcore}, CFA~\cite{cfa}, DRAEM~\cite{draem}, SimpleNet~\cite{simplenet}, and UniAD~\cite{uniad}. Here, most methods are run with the publicly available \href{https://github.com/openvinotoolkit/anomalib}{Anomalib} except for  DRAEM~\cite{draem}, SimpleNet~\cite{simplenet}, and UniAD~\cite{uniad} using official code.

\subsection{Comparisons with State-of-the-Arts}
\textbf{Main Results.} We report the results of image-level classification and pixel-level segmentation on three industry AD datasets (MVTec, BTAD and VisA) and compare our OneNIP with state-of-the-art methods in Tab.~\ref{tab:comsota}. Some important observations are summarised as follows: 

Most state-of-the-art methods suffer from a significant performance drop in both image-level classification and pixel-level segmentation when extending one-model-one-class setting to a one-model-all-classes one, which is also consistent with observations in UniAD. For example, state-of-the-art SimpleNet~\cite{simplenet} drops about 21.4\% (from 99.6\% to 78.2\%) in I-ROC and 17.1\% (from 98.1\% to 81.0\%) in P-ROC, respectively; 
Our method beats all competitors and outperforms the state-of-the-art UniAD by a large margin for pixel-level anomaly segmentation on all three datasets, e.g., from 44.7\% to 63.7\% on MVTec, from 50.9\% to 56.8\% on BTAD, and 33.6\% to 43.3\% on VisA; 
Some methods are not robust to different application scenarios while our method consistently outperforms state-of-the-art methods. For example, DRAEM achieves 49.6\% P-PR on MVTec, but only 12.3\% on BTAD and 15.1\% on VisA; 
For image-level anomaly classification, our method also surpasses UniAD in most cases, e.g., improving I-ROC performance from 96.5\% to 97.9\% on MVTec, and 90.8\% to 92.5\% on VisA. 

\begin{table*}[t]
\setlength\tabcolsep{5pt}
\centering
\small
\caption{\small\textbf{Image-level anomaly classification and pixel-level anomaly segmentation comparisons with ROC/PR metric on MVTec, BTAD and VisA.} All methods are evaluated under the unified setting. The best and second-best results are highlighted in~\textcolor{red}{red} and \textcolor{blue}{blue}, respectively. Note that the results are averaged over multiple categories and the full results of each category are presented in supplementary material.}
\label{tab:comsota}
\vspace{-5pt}
\resizebox{1.0\textwidth}{!}{
\begin{tabular}{c|c|c cc cc| cc |cc}
\toprule
    &   &\multicolumn{5}{c|}{\textbf{Embedding-based}}  & \multicolumn{2}{c|}{\textbf{Discriminator-based}} & \multicolumn{2}{c}{\textbf{Reconstruction-based}}   \\ 
    \cmidrule{3-11}
  \multirow{-2}{*}{\textbf{Datasets}} &\multirow{-2}{*}{\textbf{Metric}$\uparrow$}	&\footnotesize{CS-Flow\cite{csflow}}	&\footnotesize{PaDiM\cite{padim}}  	&\footnotesize{DFM\cite{dfm}} 	&\footnotesize{PatchCore\cite{patchcore}} 	&\footnotesize{CFA\cite{cfa}}	&\footnotesize{DRAEM\cite{draem}}	&\footnotesize{SimpleNet\cite{simplenet}}  	&\footnotesize{UniAD\cite{uniad}} 	&\footnotesize{\textbf{OneNIP}} \\
\midrule
&I-ROC/PR	&81.4 / 90.2	&87.5 / 92.8	&69.7 / 89.8	&89.8 / 96.3	&80.4 / 91.0	&91.4 / 95.3	&78.2 / 90.0	 &\blue{96.5} / \blue{98.9}	&\red{97.9} / \red{99.3}\\
 \multirow{-2}{*}{\textbf{MVTec}\cite{mvtec}} 
&P-ROC/PR	&93.8 / 33.8	&95.5 / 37.8	&96.5 / 42.4 	&96.4 / \blue{50.1}	&90.7 / 37.1	&85.2 / 49.6	&81.0 / 24.8	&\blue{96.8} / 44.7	&\red{97.9} / \red{63.7}  \\
 \midrule
&I-ROC/PR	&91.8 / 96.3 	&\red{95.7} / 97.4	&68.8 / 82.8 	&89.2 / 96.4	&87.5 / 87.7	&84.7 / 95.0 	&90.3 / 95.0	&92.2 / \blue{97.9}	&\blue{92.6} /  \red{98.5}\\
 \multirow{-2}{*}{\textbf{BTAD}\cite{btad}} 
 &P-ROC/PR	&95.9 / 34.6	&96.7 / 48.7	&96.3 / 48.0	&96.3 / 48.4 	&95.6 / 40.4	&74.2 / 12.3	&78.8 / 36.2	&\blue{97.1} / \blue{50.9}	&\red{97.4} / \red{56.8}\\
 \midrule
&I-ROC/PR	&75.8 / 80.0	&78.1 / 78.3	&51.6 / 77.8	&90.3 / 92.0 	&69.0 / 73.8 	&81.8 / 85.8 	&89.2 / 92.2	&\blue{90.8} / \blue{93.0}	&\red{92.5} / \red{94.5} \\
 \multirow{-2}{*}{\textbf{VisA}\cite{visa}} 
&P-ROC/PR	&95.6 / 18.6	&95.9 / 17.1	&96.5 / 25.2	&96.8 / 38.2	&91.4 / 16.8	&78.1 / 15.1	&95.3 / 33.1	&\blue{98.4} / \blue{33.6}	&\red{98.7} / \red{43.3} \\
\bottomrule
\end{tabular}}
\vspace{-5pt}
\end{table*}

Furthermore, we also compared the trend of testing metrics (I-ROC, P-ROC and P-PR) for UniAD and our OneNIP with the number of training epochs increased, as shown in \cref{fig:uniadvsonenip-b}. It can be observed that our method only requires significantly fewer epochs to achieve the same performance as UniAD, especially for P-PR. This reveals that the introduction of a normal image prompt and supervised refiner indeed accelerates the convergence of the reconstruction model.

\begin{wraptable}{r}{0.6\textwidth}
\vspace{-35pt}
\centering
\small
\caption{\small{\textbf{Comparisons with state-of-the-art UniAD on a more complex data distribution (one model for multiple datasets).} 
\vspace{5pt}
}}
\label{tab:allad}
\resizebox{0.6\textwidth}{!}{
    \setlength\tabcolsep{0.9pt}
\begin{tabular}{c|c|c|cc}
\toprule
 Datasets &\#Classes &Metric $\uparrow$  &UniAD~\cite{uniad}  &OneNIP \\
 \hline
 \multirow{2}{*}{MVTec\cite{mvtec}} &\multirow{2}{*}{15} 
&{I-ROC/PR}  &94.8/98.0 &\red{97.1}/\red{99.0}\\
& &{P-ROC/PR}  &96.2/42.1 &\red{97.6}/\red{61.1}\\
\midrule
 \multirow{2}{*}{BTAD\cite{btad}}  &\multirow{2}{*}{3} 
&{I-ROC/PR}  &92.0/97.1&\red{92.0}/\red{97.5}\\
&&{P-ROC/PR}  &97.1/48.0&\red{97.9}/\red{59.0}\\
\midrule
 \multirow{2}{*}{VisA\cite{visa}}  &\multirow{2}{*}{12} 
&{I-ROC/PR} &89.9/92.4 &\red{91.9}/\red{93.9}\\
&&{P-ROC/PR}  &98.3/33.2 &\red{98.6}/\red{40.6}\\
\midrule
 \multirow{2}{*}{\textbf{All}}   &\multirow{2}{*}{30} 
&{I-ROC/PR} &92.6/95.7  &\red{94.5}/\red{96.8}\\
&&{P-ROC/PR} &97.1/39.1  &\red{98.0}/\red{52.4}\\
\bottomrule
\end{tabular}}
\vspace{-10pt}
\end{wraptable}

\noindent\textbf{Results on More Complex Distribution.}
In Tab.~\ref{tab:comsota}, we train a unified anomaly detection model for each dataset following the previous UniAD. To further demonstrate the effectiveness of the proposed OneNIP when facing a more complex data distribution, we merge MVTec, BTAD, and VisA into a larger scale and more categories dataset, then train UniAD and OneNIP on the merging dataset. We report the image-level classification and pixel-level segmentation results including the average over all 30 categories, and the mean results of each dataset in Tab.~\ref{tab:allad}.

Our OneNIP still significantly outperforms the state-of-the-art UniAD in both image-level classification (94.5\% vs. 92.6\% in I-ROC) and pixel-level segmentation (52.4\% vs. 39.1\% in P-PR) when evaluating on a more complex data distribution~(i.e., one model for multiple datasets). Furthermore, there is no significant performance drop from the unified case (one model for multi-class) in Tab.~\ref{tab:comsota} to a more unified case (one model for multi-dataset) in Tab.~\ref{tab:allad}, while most existing methods suffer from a significant performance drop when they are extended to complex distributions (i.e., one model for all classes).

\begin{wraptable}{r}{0.6\textwidth}
\vspace{-35pt}
\centering
\small
\caption{\small{\textbf{Results comparisons of OneNIP with different resolutions on MVTec, BTAD and VisA.}}}
\label{tab:inputres}
\vspace{5pt}
\resizebox{0.6\textwidth}{!}{
    \setlength\tabcolsep{0.9pt}
\begin{tabular}{c|c|ccc}
\toprule
 Datasets &Metric $\uparrow$  &~~224$\times$224~~  &~~256$\times$256~~  &~~320$\times$320~~\\
 \hline
 \multirow{2}{*}{MVTec\cite{mvtec}} 
&\small{I-ROC/PR}  &\red{97.9}/\red{99.3} &97.6/99.2 & \red{97.9}/\red{99.3}\\
&\small{P-ROC/PR}  &\red{97.9}/63.7 &97.8/64.7 & \red{97.9}/\red{65.9}\\
\midrule
 \multirow{2}{*}{BTAD\cite{btad}}  
&\small{I-ROC/PR}  &92.6/98.5 &94.9/\red{99.0}	&\red{95.3}/98.9\\
&\small{P-ROC/PR}  &97.4/56.8 &97.6/57.0	&\red{97.8}/\red{57.6}\\
\midrule
 \multirow{2}{*}{VisA\cite{visa}}  
&\small{I-ROC/PR} &92.5/94.5 &93.3/94.3 &\red{94.2}/\red{95.7}
\\
&\small{P-ROC/PR} &98.7/43.3 &98.8/44.1 &\red{98.8}/\red{46.1}\\
\bottomrule
\end{tabular}}
\vspace{-10pt}
\end{wraptable}

\noindent\textbf{Results on Different Resolutions.}
We conduct OneNIP with varying input resolutions considering different defect area distributions on different datasets, and the results are reported in Tab.~\ref{tab:inputres}. For pixel-level anomaly segmentation, the performance tends to consistently improve when the input resolution increases in a certain range (i.e., from 224$\times$224 to 320$\times$320). For image-level anomaly classification, accuracy can be significantly boosted when increasing input resolution on BTAD and VisA, while almost constant on MVTec. This is not surprising as we know that anomaly regions are typically smaller on BTAD and VisA compared to MVTec. The low resolution makes it challenging for pre-trained models to capture anomaly characters, thus resulting in difficulties in small anomaly detection.

 \begin{figure*}[t]
  \centering
  \includegraphics[width= 0.065\textwidth]{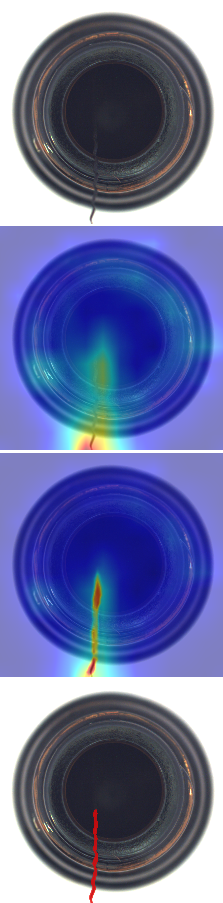}
  \hspace{-5.5pt}
  \includegraphics[width= 0.065\textwidth]{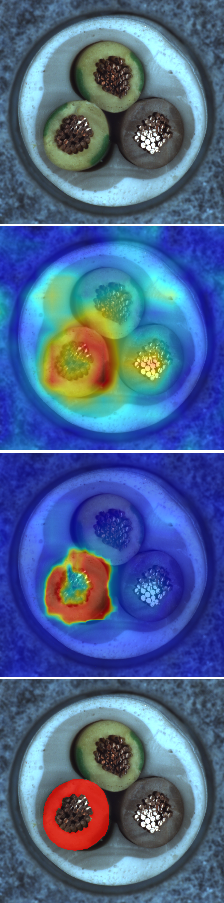}
  \hspace{-5.5pt}
  \includegraphics[width= 0.065\textwidth]{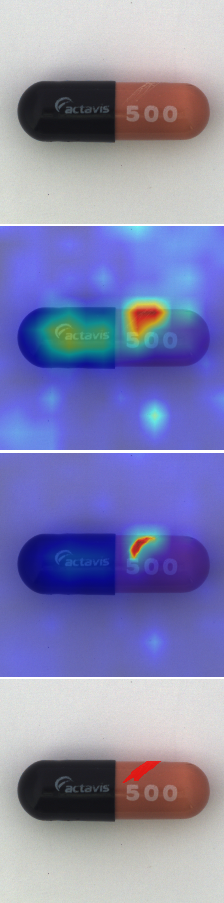}
  \hspace{-5.5pt}
  \includegraphics[width= 0.065\textwidth]{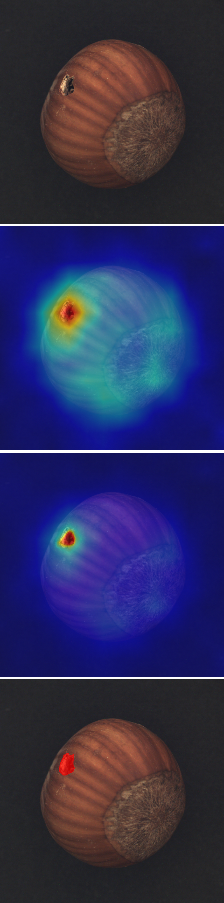}
  \hspace{-5.5pt}
  \includegraphics[width= 0.065\textwidth]{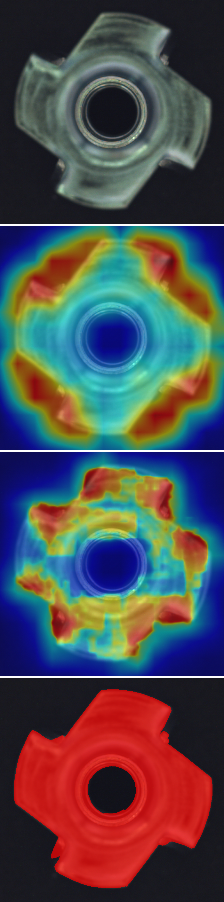}
  \hspace{-5.5pt}
  \includegraphics[width= 0.065\textwidth]{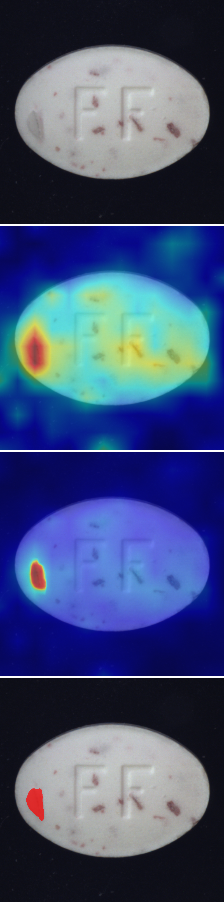}
  \hspace{-5.5pt}
  \includegraphics[width= 0.065\textwidth]{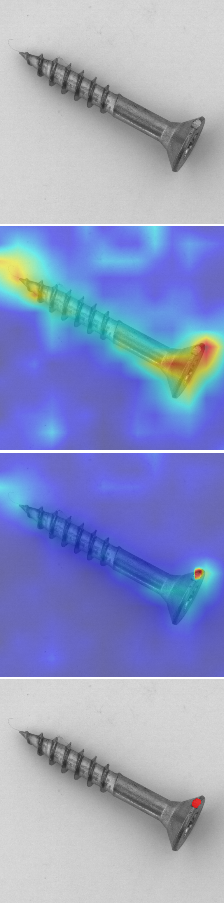}
  \hspace{-5.5pt}
  \includegraphics[width= 0.065\textwidth]{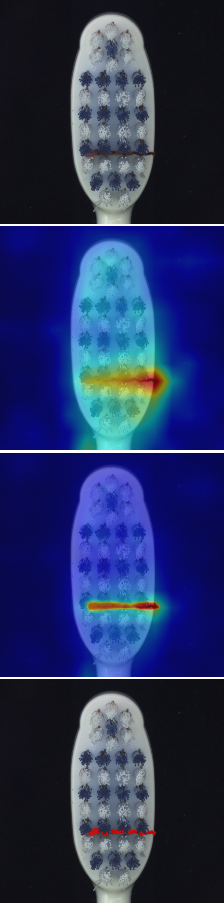}
  \hspace{-5.5pt}
  \includegraphics[width= 0.065\textwidth]{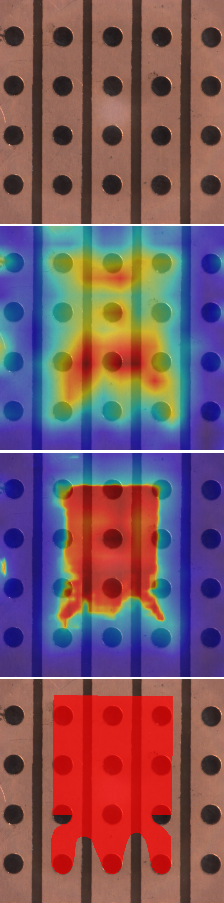}
  \hspace{-5.5pt}
  \includegraphics[width= 0.065\textwidth]{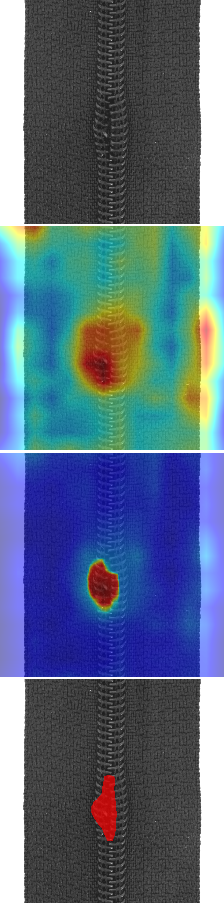}
  \hspace{-5.5pt}
  \includegraphics[width= 0.065\textwidth]{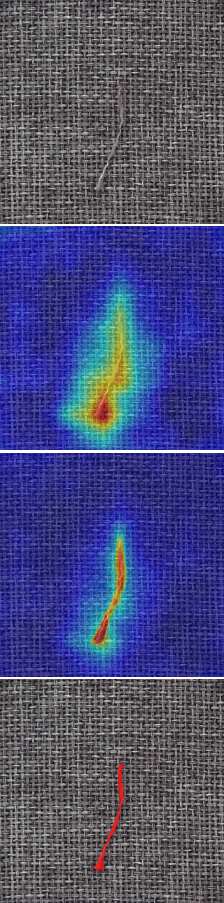}
  \hspace{-5.5pt}
  \includegraphics[width= 0.065\textwidth]{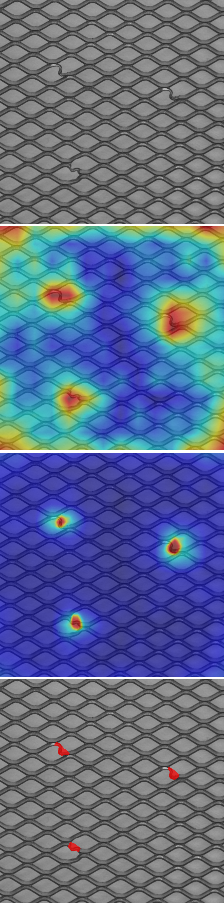}
  \hspace{-5.5pt}
  \includegraphics[width= 0.065\textwidth]{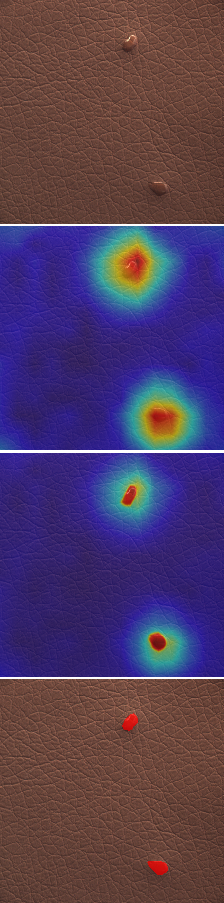}
  \hspace{-5.5pt}
  \includegraphics[width= 0.065\textwidth]{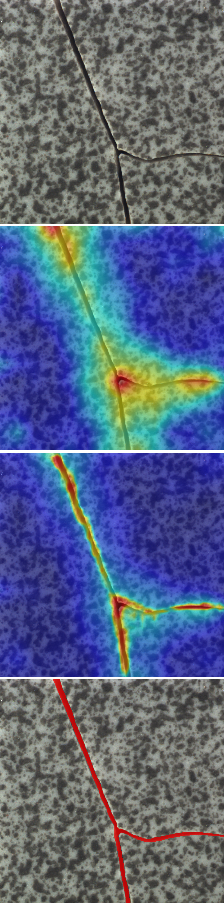}
  \hspace{-5.5pt}
  \includegraphics[width= 0.065\textwidth]{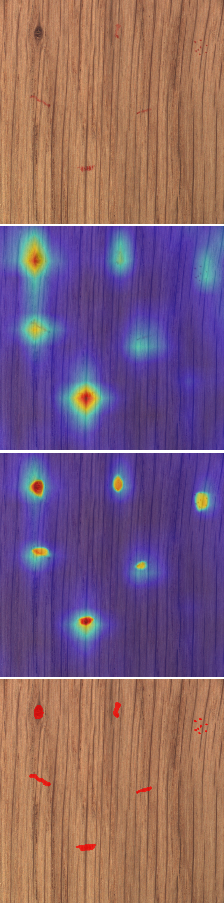}\\
  \includegraphics[height= 0.205\textwidth]{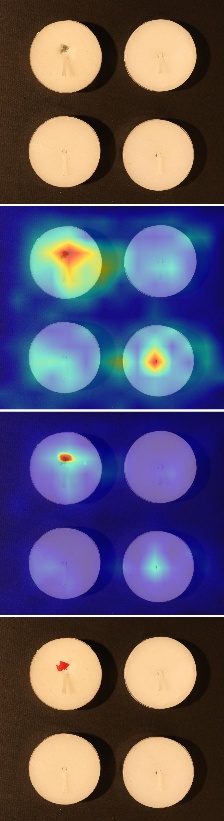}
  \hspace{-5.5pt}
  \includegraphics[height= 0.205\textwidth]{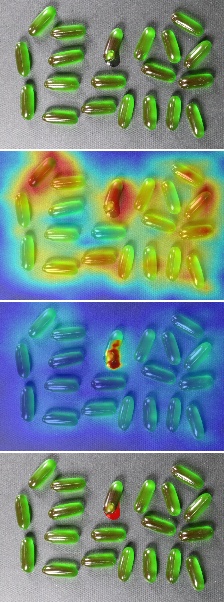}
  \hspace{-5.5pt}
  \includegraphics[height= 0.205\textwidth]{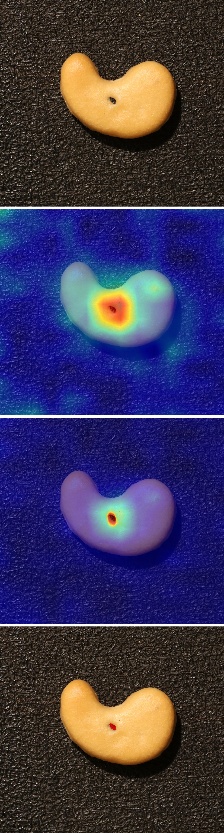}
  \hspace{-5.5pt}
  \includegraphics[height= 0.205\textwidth]{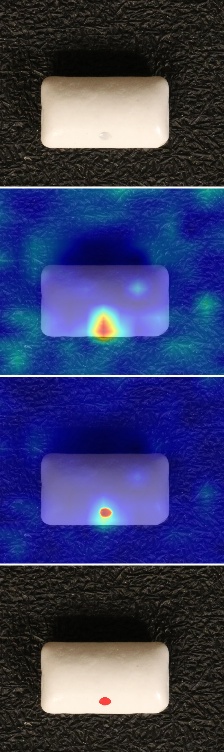}
  \hspace{-5.5pt}
  \includegraphics[height= 0.205\textwidth]{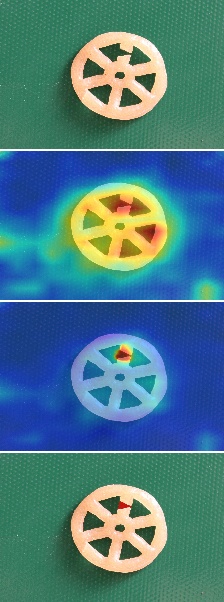}
  \hspace{-5.5pt}
  \includegraphics[height= 0.205\textwidth]{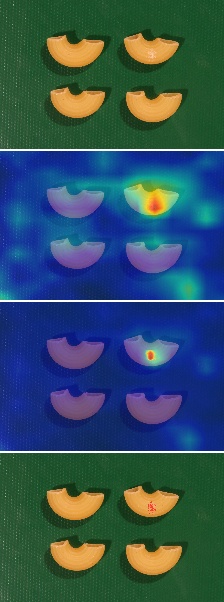}
  \hspace{-5.5pt}
  \includegraphics[height= 0.205\textwidth]{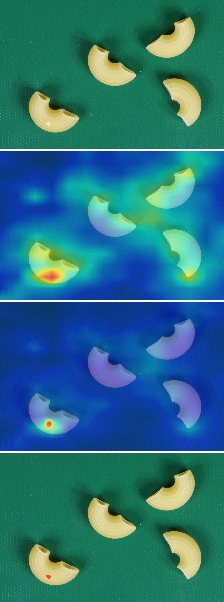}
  \hspace{-5.5pt}
  \includegraphics[height= 0.205\textwidth]{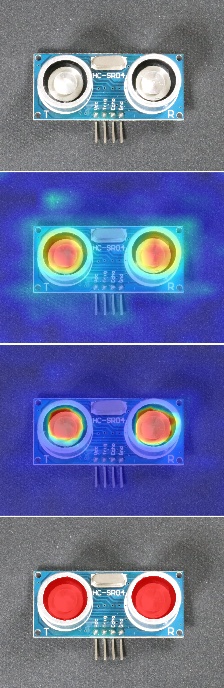}
  \hspace{-5.5pt}
  \includegraphics[height= 0.205\textwidth]{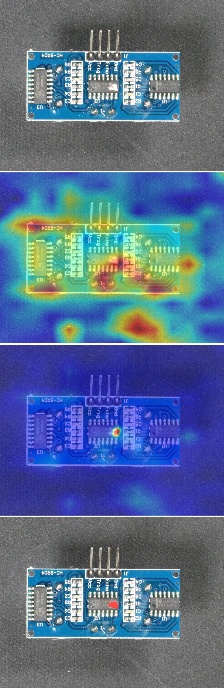}
  \hspace{-5.5pt}
  \includegraphics[height= 0.205\textwidth]{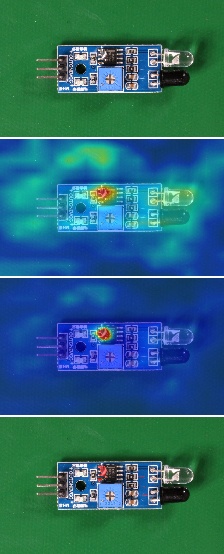}
  \hspace{-5.5pt}
  \includegraphics[height= 0.205\textwidth]{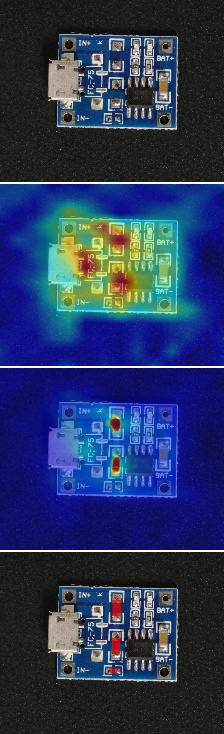}
  \hspace{-5.5pt}
  \includegraphics[height= 0.205\textwidth]{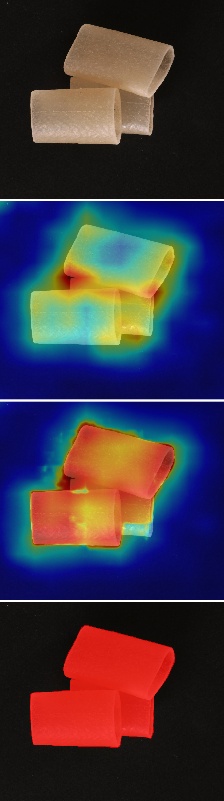}
  \hspace{-5.5pt}
  \includegraphics[height= 0.205\textwidth]{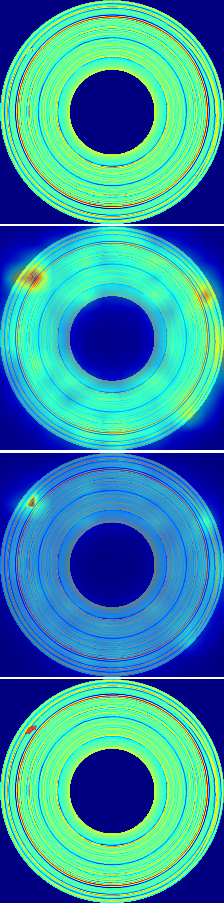}
  \hspace{-5.5pt}
  \includegraphics[height= 0.205\textwidth]{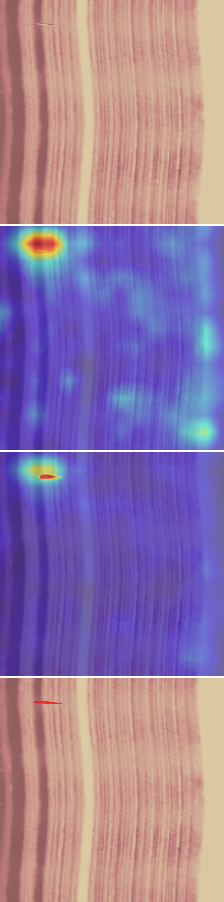}
  \hspace{-5.5pt}
  \includegraphics[height= 0.205\textwidth]{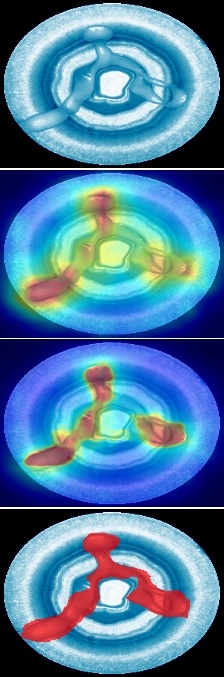} \\
  \caption{\small{Qualitative comparisons of \href{https://github.com/zhiyuanyou/UniAD}{UniAD} (second and sixth rows) and our OneNIP (third and seventh rows) on MVTec~(15 classes), BTAD~(3 classes) and VisA~(12 classes).} Here, the first and fifth rows are original testing images, and the fourth and eighth rows are their corresponding anomaly masks highlighted with red color.}
  \label{fig:quacom}
  \vspace{-5pt}
 \end{figure*}

\noindent\textbf{Qualitative Comparisons.}
We present representative examples to qualitatively compare UniAD and our OneNIP for each object on MVTec, BTAD and VisA in Fig.~\ref{fig:quacom}. It can be observed that both UniAD and OneNIP are able to recognize anomalies at image level, but OneNIP often does more precise segmentation at pixel level.

\subsection{Ablation Studies}\label{subsec:ablation}

To verify the effectiveness of all proposed components and the effects of hyperparameters, we implement extensive ablation studies on MVTec with a unified setting as shown in Tab.~\ref{tab:as}.

\definecolor{azure}{rgb}{0.0, 0.3, 1.0}
\begin{table*}[t]
\centering
\caption{\small{
    \textbf{Ablation studies on MVTec}. 
    Default settings are marked in \textcolor{azure}{blue}. 
}}
\vspace{-5pt}
\label{tab:as}
\begin{minipage}[t]{0.55\textwidth}
\centering
\small{(a) Prompt strategy in Reconstruction, Restoration, and Refiner}\label{tab:asa}
\setlength\tabcolsep{0.5pt}
\begin{tabular}{c|ccc|cccc}
\toprule
No. &Prompt & Res. &Ref. &I-ROC &P-ROC  &I-PR &P-PR \\
\midrule
\gray{0} &\ding{55}	&\ding{55}	&\ding{55}	&96.5	&96.8	&98.9	&44.7 \\
\gray{1} &static	&\ding{55}	&\ding{55}	&96.8	&97.0	&98.9	&45.8\\
\gray{2} &dynamic	&\ding{55}	&\ding{55}	&97.5	&97.1	&99.2	&46.0\\
\gray{3} &\ding{55}	&\ding{51}	&\ding{55}	&96.7	&97.0	&98.9	&46.5\\
\gray{4} &dynamic	&\ding{51}	&\ding{55}	&97.4	&97.3	&99.1	&48.4\\
\gray{5} &\textcolor{azure}{dynamic}	&\textcolor{azure}{\ding{51}}	&\textcolor{azure}{\ding{51}}	&\textbf{97.9}	&\textbf{97.9}	&\textbf{99.3}	&\textbf{63.7}\\
\bottomrule
\end{tabular}
\end{minipage}
\hspace{5pt}
\begin{minipage}[t]{0.41\textwidth}
\centering
\small{(b) Effects of the number of Encoder, and Decoder}\label{tab:asb}
\setlength\tabcolsep{0.5pt}
\begin{tabular}{cc|cccc}
\toprule
Enc &Dec  &I-ROC &P-ROC  &I-PR &P-PR \\
\midrule
1	&1	&94.8	&97.0	&97.9	&56.0\\
2	&2	&96.7	&97.4	&98.9	&59.4\\
\textcolor{azure}{4}	&\textcolor{azure}{4}	&97.9	&97.9	&99.3	&63.7\\
6	&6	&\textbf{98.1}	&\textbf{98.0}	&\textbf{99.4}	&\textbf{64.6}\\
2	&4	&97.0	&97.6	&99.0	&61.2\\
4	&2	&97.1	&97.6	&99.0	&62.1\\
\bottomrule
\end{tabular}
\end{minipage}

\vspace{5pt}
\begin{minipage}[t]{0.34\textwidth}
\centering
\small{(c)  Effects of weight $\alpha$ 
}
\setlength\tabcolsep{0.5pt}
\begin{tabular}{c|cccc}
\toprule
$\alpha$  &I-ROC &P-ROC  &I-PR &P-PR \\
\midrule
0.00 &97.6	&97.3	&99.2	&48.3\\
0.25 &97.8	&97.7	&99.3	&59.3 \\
\textcolor{azure}{0.50} &97.9	&97.9	&99.3	&63.7\\
1.00 &96.7	&96.7	&98.9	&63.7\\
\bottomrule
\end{tabular}
\end{minipage}
\hspace{5pt}
\begin{minipage}[t]{0.60\textwidth}
\centering
\small{(d) Different prompt modes of the same category 
}
\setlength\tabcolsep{0.5pt}
\begin{tabular}{lc|llll}
\toprule
Train &Test &I-ROC &P-ROC  &I-PR &P-PR \\
\midrule
\multirow{2}{*}{\textcolor{azure}{rand}} &rand 
&97.85\begin{scriptsize}$\pm$0.01\end{scriptsize} 	&97.86\begin{scriptsize}$\pm$0.00\end{scriptsize} 	&99.27\begin{scriptsize}$\pm$0.01\end{scriptsize}  	&63.71\begin{scriptsize}$\pm$0.01\end{scriptsize}  	\\

&\textcolor{azure}{fixed}  
&97.85\begin{scriptsize}$\pm$0.02\end{scriptsize}  	&97.86\begin{scriptsize}$\pm$0.00\end{scriptsize} 		&99.27\begin{scriptsize}$\pm$0.01\end{scriptsize} 	&63.71\begin{scriptsize}$\pm$0.02\end{scriptsize}  	\\
\hline
\multirow{2}{*}{fixed}  &fixed   &97.91	&97.86	&99.30	&63.66\\
&rand 
&96.05\begin{scriptsize}$\pm$0.24\end{scriptsize}	&97.49\begin{scriptsize}$\pm$0.03\end{scriptsize}	&98.34\begin{scriptsize}$\pm$0.19\end{scriptsize}	&60.65\begin{scriptsize}$\pm$0.18\end{scriptsize}\\
\bottomrule
\end{tabular}
\end{minipage}
\vspace{-5pt}
\end{table*}

\noindent\textbf{Static or Dynamic Prompt in Reconstruction.} We first simply replace the learned query embedding with a normal image prompt feature in the LQD of UniAD, which brings 1.1 points improvement in P-PR for anomaly segmentation and also improves image-level anomaly classification in I-ROC and P-ROC (Lines 0 and 1 in Tab.~\ref{tab:as}a). Further, there is a significant improvement in both image-level classification and pixel-level segmentation when we take the static prompt as an initial value and dynamically update the prompt and target feature in our bidirectional decoder (Lines 0 and 2 in Tab.~\ref{tab:as}a). This demonstrates that the dynamic prompt manner takes an important role in unsupervised feature reconstruction.

\noindent\textbf{Effectivness of Restoration.} To enhance the guidance of prompt in unsupervised reconstruction, we introduce synthesized anomaly images and restore their features to normal ones and thus form an unsupervised restoration stream. In fact, this is what we expect at the inference stage. We can see that the restoration stream is effective in improving pixel-level anomaly segmentation (from 46.0\% to 48.4\% in P-PR, Lines 2 and 4 in Tab.~\ref{tab:as}a). For fair comparisons, we directly introduce the restoration stream into UniAD without normal prompt, but the corresponding improvement is less than ours (Lines 3 and 4 in Tab.~\ref{tab:as}a). This further implies the importance of normal prompts in the restoration stream.

\noindent\textbf{Effectivness of Refiner.} We improve anomaly localization by regressing reconstruction errors from low to high resolution with a supervised refiner. It is simple and lightweight but greatly boosts pixel-level anomaly segmentation (from 48.4\% to 63.7\% in P-PR, Lines 4 and 5 in Tab.~\ref{tab:as}a). This can be attributed to two facts: reconstruction errors themselves can roughly localize anomalies, and pseudo anomalies carry accurate pixel-level masks.

\noindent\textbf{Effects of Hyper-parameter.} 
We carefully study the effects of some hyper-parameters, such as the number of encoder and decoder, coefficient $\alpha$ between reconstruction and refiner, and prompt mode in training and testing as shown in Tab.~\ref{tab:as}b, c, and d. It can be seen that more encoders and decoders are helpful for better performance. Furthermore, it is important to choose a reasonable $\alpha$, which will be dependent on the refiner when the coefficient is too large, and on the reconstruction when it's too small. To demonstrate the robustness of our method to different normal prompts of the same category, we compare different prompt modes, i.e., random and fixed in training and testing, and report averaged testing metric and standard deviation based on 10 random seeds in Tab.~\ref{tab:as}d.
It can be seen that our method is robust for different normal prompts of the same category when training in a random mode. However, incorrect image prompts will significantly decrease the performance. For example, if one MetaNut image as the prompt for testing the Screw images, the corresponding I-ROC drops from 91.4\% to 67.3\%, and P-PR drops from 39.8\% to 2.3\%. Furthermore, it will weaken performance if training the model with a fixed image prompt while testing in a random manner. This performance degradation mainly happens for large positional changes, such as Srew on MVTec, and the performance of most categories is maintained because the geometric appearance of MVTec for most categories is roughly aligned.

 \section{Conclusion}\label{sec:conclusion}

In this paper, we propose a simple yet effective anomaly detection framework that learns to detect anomalies with a normal image prompt. To adequately leverage the prompt information in unsupervised feature reconstruction, we first propose a bidirectional decoder to dynamically update the prompt and target features and promote their interaction. To further enhance the guidance of the prompt, we introduce pseudo-anomalous images and propose a restoration stream that restores these pseudo-anomalous features to the corresponding normal ones. Furthermore, we propose a lightweight refiner that regresses the reconstruction errors for both real normal and pseudo-anomalous samples from low to high resolution in a supervised manner, which greatly boosts anomaly segmentation performance.

\noindent\textbf{Limitation:} In our OneIP, the proposed restoration stream introduces additional training costs, although it can be completely removed at inference. Furthermore, the bidirectional decoder and supervised refiner are only simply designed, leaving ample room for improvement.


%
%
\bibliographystyle{splncs04}
\bibliography{onenip}
\clearpage
\setcounter{page}{1}
\appendix

\section{Implementation Details}
For fair comparisons, we maintain the same hyper-parameters as in UniAD~\cite{uniad}. All input images are resized to 224$\times$224 resolution for all methods both training phase and inference time. The 4 staged features extracted from stages 1 to 4 of EfficientNet-b4 are first resized to a spatial size of 14$\times$14 and then concatenated together to finally form a 272-channel feature map. For unsupervised reconstruction or restoration, the layer numbers of the encoder and decoder are set to 4 to balance performance and computation costs. For supervised refiner, we employ two transposed convolutional blocks, and the channels of each convolution block are set to 128. For synthesized anomaly generation, we employ CutPaste~\cite{cutpaste} and DRAEM~\cite{draem} with a probability of 0.5. The loss weight $\lambda$ is set to 0.5.

The model is trained with a total of 1000 epochs on 8 Tesla V100 GPUs with batch size 64. AdamW optimizer with weight decay $1\times10^{-4}$ is used. The learning rate is $1\times10^{-4}$ initially and dropped by 0.1 after 800 epochs. We conduct experiments based on the open-source framework PyTorch and NVIDIA V100 GPU. We use the official codes for DRAEM~\cite{draem}, SimpleNet~\cite{simplenet} and UniAD~\cite{uniad}, and the publicly available \href{https://github.com/openvinotoolkit/anomalib}{Anomalib} for other methods.

\section{Industry Anomaly Detection Benchmarks}
Following previous works, we comprehensively evaluate our method on three industry anomaly detection benchmarks, {MVTec}~\cite{mvtec}, {BTAD}~\cite{btad}, and VisA~\cite{visa}.

\noindent\textbf{MVTec}~\cite{mvtec} is a highly popular dataset used for industrial anomaly detection. It encompasses 15 categories (10 objects and 5 textures) from real-world manufacturing. The whole dataset is split into training and testing sets. The training set includes 3,629 normal images, and the testing set contains 1,258 anomaly images and 467 normal images. All anomaly images are annotated by pixel-level mask, which is very convenient for pixel-level evaluation.

\noindent\textbf{BTAD}~\cite{btad} is another real-world industrial anomaly detection dataset. It contains a total of 2,830 images, showcasing 3 industrial products with body and surface defects. The training set comprises 1,799 normal images while the testing set consists of 290 anomaly images and 451 normal images. Similar to the MVTec, pixel-wise annotations are given for anomaly images in the testing set.

\noindent\textbf{VisA}~\cite{visa} is currently a larger and more challenging anomaly detection dataset. This dataset contains 12 objects spanning 3 domains, complex structures, multiple instances, and multiple anomaly classes. There are 10,821 high-resolution color images with 9,621 normal (8,659 for training and 962 for testing) and 1,200 anomaly images (all for testing) carrying both image- and pixel-level annotations.

\section{Complete Multi-class Anomaly Detection Results}\label{sec:comres}
In our main paper, we reported only the averaged results of all categories for each dataset. Here, we provide a more comprehensive report in Tabs.~\ref{tab:ad_pixel} and \ref{tab:ad_image}, detailing both image-level anomaly classification and pixel-level anomaly segmentation for each category on MVTec, BTAD, and VisA.

\begin{table*}[t]
\setlength\tabcolsep{5pt}
\centering
\small
\caption{\small\textbf{Pixel-level anomaly segmentation comparisons with ROC/PR on MVTec, BTAD and VisA.} All methods are evaluated under the unified setting. The best and second-best results are highlighted in~\textcolor{red}{red} and \textcolor{blue}{blue}, respectively.}
\label{tab:ad_pixel}
\vspace{-5pt}
\resizebox{1.0\textwidth}{!}{
\begin{tabular}{c|c|c cc cc| cc |cc}
\toprule
 &   &\multicolumn{5}{c|}{\textbf{Embedding-based}}  & \multicolumn{2}{c|}{\textbf{Discriminator-based}} & \multicolumn{2}{c}{\textbf{Reconstruction-based}}   \\ 
    \cmidrule{3-11}
\multirow{-2}{*}{Datasets} &\multirow{-2}{*}{Catergories}	&\footnotesize{CS-Flow\cite{csflow}}	&\footnotesize{PaDiM\cite{padim}}  	&\footnotesize{DFM\cite{dfm}} 	&\footnotesize{PatchCore\cite{patchcore}} 	&\footnotesize{CFA\cite{cfa}}	&\footnotesize{DRAEM\cite{draem}}	&\footnotesize{SimpleNet\cite{simplenet}}  	&\footnotesize{UniAD\cite{uniad}} 	&\footnotesize{\textbf{OneNIP}} \\
\cmidrule{1-11}
&Bottle & 97.5 / 61.7 & 97.7 / 64.2 & 97.1 / 58.2 & 97.3 / 72.1 & 94.4 / 49.1 & 96.1 / \textcolor{blue}{75.9} & 79.8 / 21.0 & \textcolor{blue}{98.0} / 68.0 &\red{98.5}	/ \red{81.6}\\
&Cable    & 84.5 / 15.7 & 95.9 / 36.6 & 97.4 / 51.8 & \textcolor{red}{98.2} / \textcolor{blue}{56.0} & 90.8 / 22.6 & 52.4 / ~~5.7 & 82.8 / 24.4 & 97.5 / 53.1 & \red{98.2} / \red{67.5}\\
&Capsule  & 97.8 / 30.4 & \textcolor{blue}{98.1} / 31.0 & 97.1 / 30.8 & 97.5 / 36.7 & 97.7 / 39.7 & 66.5 / ~~7.7 & 89.4 / 12.5 & \textcolor{red}{98.6} / \textcolor{blue}{46.5} & \red{98.6}	/ \red{49.9}\\
&Hazelnut & 96.0 / 32.2 & 98.1 / 48.0 & 98.1 / 44.4 & \blue{98.5} / 49.6 & 97.7 / 48.5 & 97.5 / \red{73.9} & 88.5 / 20.3 & 98.1 / 53.6 &\red{98.7}	/ \blue{70.2}\\
&Metal Nut& 87.9 / 41.3 & 95.3 / 66.9 & \blue{97.8} / \red{81.4} & \red{98.6} / \blue{80.8} & 95.2 / 66.9 & 56.5 / 22.8 & 89.5 / 53.3 & 93.3 / 49.5 & 96.5	/ 74.1\\
&Pill     & 94.1 / 38.0 & 95.1 / 37.6 & \blue{97.3} / 55.9 & \red{97.7} / \red{59.0} & 94.1 / \blue{58.5} & 75.4 / 37.4 & 83.1 / 42.3 & 95.4 / 40.7  &96.0	/ 47.6\\
&Screw    & 95.2 / ~~7.0 & 97.0 / ~~9.1 & 95.8 / ~~5.9 & 96.4 / ~~8.9 & 94.9 / ~~3.6 & 97.3 / \red{55.8} & 69.9 / ~~0.6 & \red{98.4} / 24.5 &\red{98.9}	/ 39.8\\
&Toothbrush & 97.6 / 38.6 & 98.1 / 39.6 & 98.2 / 51.4 & 98.2 / 47.3 & 97.5 / 49.4 & 97.6 / \red{68.9} & 92.1 / 31.8 & \blue{98.4} / 39.7 &\red{98.8}	/ 53.4\\
&Transistor & 84.5 / 38.7 & 96.1 / 49.0 & 98.2 / 69.6 & 96.1 / 69.7 & 82.9 / 24.0 & 74.1 / 27.9 & 69.0 / 19.1 & \blue{98.3} / \blue{73.2} &\red{98.8}	/ \red{84.1} \\
&Zipper     & 96.1 / 37.4 & 92.9 / 21.6 & 95.2 / 24.5 & 95.4 / {56.9} & 85.8 / 19.5 & 96.3 / \red{63.5} & 88.4 / 25.5 & \blue{96.6} / 33.2 &\red{97.6}	/ \blue{59.1}\\
&Carpet     & 98.0 / 37.4 & \blue{98.5} / 49.2 & 98.3 / 48.3 & 97.5 / 49.4 & 92.3 / 36.3 & 94.4 / \blue{60.7} & 94.7 / 31.4 & \blue{98.5} / 51.4 & \red{99.0}	/ \red{69.5}\\
&Grid       & 92.9 / 17.4 & 86.2 / 11.5 & 93.8 / 14.2 & 87.3 / 12.0 & 53.2 / ~~0.7 & \red{98.7} / \red{54.9} & 26.6 / ~~0.3 & 96.6 / 22.2  &\blue{98.4}	/ \blue{45.6}\\
&Leather    & \blue{98.9} / 37.5 & 98.7 / 29.4 & 98.3 / 25.1 & \blue{98.9} / 41.3 & 98.8 / 43.5 & 96.2 / \blue{51.1} & 85.3 / ~~7.8 & 98.8 / 33.7 &\red{99.6}	/ \red{71.7}\\
&Tile       & 92.4 / 34.0 & 93.4 / 42.0 & 94.2 / 45.5 & \red{95.7} / 59.9 & 93.3 / 53.1 & 85.0 / \blue{68.2} & 90.1 / 56.0 & 92.1 / 44.2 &\blue{95.3}	/ \red{76.6}\\
&Wood       & \blue{93.9} / 40.1 & 92.0 / 31.1 & 90.5 / 28.6 & 93.3 / 52.1 & 91.4 / 40.6 & \blue{93.9} / \red{69.6} & 85.8 / 25.4 & 93.1 / 37.8 &\red{94.9}	/ \blue{65.3}	\\
\cmidrule{2-11}
\multirow{-17}{*}{\rotatebox{90}{\textbf{MVTec}\cite{mvtec}}}  
&\textbf{Mean}  & 93.8 / 33.8 & 95.5 / 37.8 & 96.5 / 42.4 & 96.4 / \blue{50.1} & 90.7 / 37.1 & 85.2 / 49.6 & 81.0 / 24.8 & \blue{96.8} / 44.7 &\red{97.9}	/ \red{63.7}\\
\midrule\midrule
\multirow{4}*{\rotatebox{90}{\textbf{BTAD\cite{btad}}}}
&01 & 95.0 / 41.8 & 95.9 / 43.5 & 94.3 / 38.4 & 94.3 / 44.6 & 92.8 / 26.6 & 87.8 / 14.4 & 90.3 / 30.2 & \blue{97.0} / \blue{53.4}  &\red{97.3}	/ \red{58.7}\\
&02 & 93.5 / 33.1 & 94.6 / 45.7 & 94.9 / \red{57.6} & \blue{95.0} / 50.4 & 94.7 / \blue{52.2} & 41.3 / ~~4.1  & 48.9 / 38.1 & 94.8 / 42.9 &\red{95.2}	/ 46.9\\
&03 & 99.4 / 28.9 & \blue{99.7} / \blue{56.8} & 99.6 / 47.9 & 99.6 / 50.2 & 99.4 / 42.3 & 93.4 / 18.3 & 97.2 / 40.4 & 99.6 / 56.4 &\red{99.8}	/ \red{64.7}\\ 
\cmidrule{2-11}
 &\textbf{Mean}  & 95.9 / 34.6 & 96.7 / 48.7 & 96.3 / 48.0 & 96.3 / 48.4 & 95.6 / 40.4 & 74.2 / 12.3 & 78.8 / 36.2 & \blue{97.1} / \blue{50.9} &\red{97.4} / \red{56.8}\\

\midrule\midrule
\multirow{13}*{\rotatebox{90}{\textbf{VisA}\cite{visa}}}
&Candle      & 97.1 / 11.6 & 97.5 / ~~8.7  & 97.9 / ~~8.0  & 95.6 / \red{45.1}   & 87.1 / ~~0.7  &  92.3 / 12.9   & 97.7 / ~~9.4  &\blue{99.1} / 20.4	&\red{99.2} / \blue{33.7}\\
&Capsules    & 86.0 / ~~7.0& 89.6 / ~~3.4  & 93.5 / 13.8   & \blue{98.0} / 13.0   & 80.6 / ~~1.0  &  67.4 / 16.3   & 94.6 / 44.9   &97.9	/ \blue{47.4}	&\red{98.4}	/ \red{55.6}\\
&Cashew      & 96.7 / 35.2 & 97.7 / 42.0   & \blue{99.2} / \red{77.1}   & 99.1 / 72.0   & 97.7 / 54.2   &  62.6 / ~~2.5  & \red{99.4} / 65.9   &99.0	/ 50.1	&\blue{99.2}	/ \blue{74.6}\\
&Chewinggum  & \blue{99.0} / 37.1 & 96.3 / 25.1   & 97.5 / 28.6   & 98.1 / 23.4   & 96.1 / 14.6   &  94.2 / 49.7   & 97.0 / 19.5   &\red{99.1}	/ \blue{57.5}	&\red{99.1}	/ \red{61.1}\\
&Fryum       & 94.2 / 22.3 & 96.9 / 42.4   & \blue{97.6} / \red{50.1}   & 92.4 / 42.2   & 93.9 / 24.1   &  74.8 / 20.0   & 93.5 / 47.3   &\red{97.7}	/ 48.2	&\red{97.7}	/ \blue{49.5}\\
&Macaroni1   & 95.3 / ~~2.5& 97.1 / ~~0.9  & 94.4 / ~~0.8  & 97.9 / ~~5.1  & 89.0 / ~~0.2  &  80.7 / \blue{12.9}   & 95.4 / ~~1.5  &\blue{99.1}	/ ~7.6	&\red{99.2}	/ \red{21.3}\\
&Macaroni2   & 92.2 / ~~0.2& 91.9 / ~~0.2  & 91.8 / ~~0.3  & 85.9 / ~~0.2  & 81.7 / ~~0.1  &  82.6 / ~~\blue{5.6}  & 83.8 / ~~0.2  &\blue{97.6}	/ ~3.1	&\red{97.9} / ~~\red{7.6}\\
&Pcb1        & 98.0 / 41.1 & 97.8 / 15.4   & 98.7 / 25.0   & \red{99.7} / \red{91.8}   & 97.8 / 41.6   &  69.7 / ~~7.5  & 99.1 / 85.6   &99.3	/ 57.4	&\blue{99.6}	/ \blue{70.0}\\
&Pcb2        & 91.6 / ~~4.6& 95.3 / ~~7.1  & 96.1 / ~~5.3  & 97.5 / 10.0   & 92.9 / ~~3.9  &  63.0 / ~~1.5  & 94.8 / \red{12.6}   &\blue{97.6}	/ ~7.7	&\red{98.1}	/ \blue{11.0}\\
&Pcb3        & 93.2 / ~~5.8& 96.1 / ~~7.3  & 94.8 / ~~7.8  & \red{99.0} / \red{43.7}   & 95.5 / ~~6.7  &  79.3 / 13.4   & 98.2 / 12.6   &98.1	/ 15.0	&\blue{98.2}	/ \blue{17.7}\\
&Pcb4        & 93.4 / ~~8.2& 95.9 / 12.1   & 96.9 / 19.6   & 97.2 / \blue{39.6}   & 87.1 / ~~5.7  &  78.1 / 12.9   & 94.5 / 28.0   &\blue{97.6}	/ 34.0	&\red{98.1}	/ \red{41.6}\\ 
&Pipe Fryum  & 98.1 / 47.5 & 98.7 / 40.0   & \blue{99.3} / {65.8}   & 99.1 / \blue{71.8}   & 97.7 / 48.7   &  92.3 / 26.6   & 95.3 / 69.6   &99.2	/ 55.1	&\red{99.5}	/ \red{76.5}\\
\cmidrule{2-11}
&\textbf{Mean} & 95.6 / 18.6 & 95.9 / 17.1 & 96.5 / 25.2   & 96.8 / \blue{38.2}   & 91.4 / 16.8   & 78.1 / 15.1    & 95.3 / 33.1  &\blue{98.4}	/ 33.6	&\red{98.7}	/ \red{43.3}\\

\bottomrule
\end{tabular}}
\vspace{-10pt}
\end{table*}
\begin{table*}[ht]
\setlength\tabcolsep{5pt}
\centering
\small
\caption{\small{\textbf{Image-level anomaly classification comparisons with ROC/PR on MVTec, BTAD and VisA.} All methods are evaluated under the unified setting. The best and second-best results are highlighted in~\textcolor{red}{red} and \textcolor{blue}{blue}, respectively.}}
\label{tab:ad_image}
\vspace{-5pt}
\resizebox{1.0\textwidth}{!}{
\begin{tabular}{c|c|ccccc|cc|cc}
\toprule
 &   &\multicolumn{5}{c|}{\textbf{Embedding-based}}  & \multicolumn{2}{c|}{\textbf{Discriminator-based}} & \multicolumn{2}{c}{\textbf{Reconstruction-based}}   \\ 
    \cmidrule{3-11}
\multirow{-2}{*}{Datasets} &\multirow{-2}{*}{Catergories}	&\footnotesize{CS-Flow\cite{csflow}}	&\footnotesize{PaDiM\cite{padim}}  	&\footnotesize{DFM\cite{dfm}} 	&\footnotesize{PatchCore\cite{patchcore}} 	&\footnotesize{CFA\cite{cfa}}	&\footnotesize{DRAEM\cite{draem}}	&\footnotesize{SimpleNet\cite{simplenet}}  	&\footnotesize{UniAD\cite{uniad}} 	&\footnotesize{\textbf{OneNIP}} \\

\midrule
&Bottle     & \red{100} / \red{100}   & 99.3 / 99.8 & 99.8 / \red{100}  & 97.6 / 99.4 & 97.5 / 99.3 & 99.6 / 99.9 & 87.1 / 95.3 & \blue{99.8} / \blue{99.9} 
& \red{100} / \red{100}\\
&Cable      & 40.2 / 61.0 & 83.9 / 87.3 & 50.0 / 80.7 & \blue{98.3} / \blue{99.1} & 68.5 / 80.6 & 63.3 / 73.0 & 76.9 / 86.0 & 95.5 / 97.3 
&\red{99.0}	/ \red{99.4}\\
&Capsule    & 84.2 / 96.5 & 77.4 / 91.0 & \blue{90.1} / \blue{97.4} & 82.6 / 96.1 & 78.7 / 94.6 & 80.9 / 95.1 & 70.6 / 92.3 & \blue{88.1} / \blue{97.1} 
&\red{91.1}	/ \red{97.8}\\
&Hazelnut   & 96.0 / 97.5 & 89.8 / 91.6 & 50.0 / 81.8 & \blue{99.9} / \red{100}  & 95.7 / 97.8 & 98.2 / \blue{98.7} & 86.9 / 93.1 & \blue{99.9} / \red{100} 
&\red{100}  / \red{100}\\
&Metal Nut  & 95.8 / 99.1 & 96.2 / 99.2 & 50.0 / 90.4 & 92.8 / 98.3 & 80.6 / 95.1 & 90.5 / 97.8 & 84.1 / 95.9 & \blue{98.9} / \blue{99.7} 
&\red{99.8}	/ \red{100}\\
&Pill       & 40.6 / 82.4 & 80.6 / 95.4 & 70.4 / 94.7 & 64.5 / 92.4 & 68.8 / 93.4 & 81.0 / 96.2 & 63.4 / 91.3 & \blue{94.0} / \blue{98.9} 
&\red{96.9}	/ \red{99.5}\\
&Screw      & 65.3 / 81.7 & 81.9 / 90.9 & 72.2 / 91.3 & 55.9 / 76.3 & 47.1 / 73.1 & \red{96.3} / \red{98.7} & 52.7 / 76.6 & 88.8 / 95.6 
&\blue{91.4}	/ \blue{96.6}\\
&Toothbrush & 75.0 / 89.9 & 71.7 / 80.2 & 84.2 / 89.5 & 83.9 / 94.1 & 75.0 / 90.7 & 89.7 / 96.1 & 85.3 / 94.2 & \red{95.8} / \red{98.3} 
&\blue{93.3}	/ \blue{97.3}\\
&Transistor & 63.2 / 60.2 & 88.9 / 85.1 & 50.0 / 70.0 & \blue{99.3} / 99.1 & 79.5 / 75.9 & 83.0 / 78.0 & 73.0 / 72.6 & \red{99.8} / \blue{99.6} 
&\red{99.8}	/ \red{99.7}\\
&Zipper     & 91.7 / 97.5 & 85.0 / 94.5 & 91.9 / 97.7 & \blue{97.2} / \blue{99.3} & 79.9 / 94.0 & \red{99.0} / \red{99.7} & 74.7 / 92.7 & 94.9 / 98.5 
&\red{99.0}	/ \red{99.7}\\
&Carpet     & 93.6 / 98.2 & 98.6 / 99.6 & 86.9 / 95.9 & 88.1 / 96.1 & 83.2 / 95.0 & 97.6 / 99.2 & 88.1 / 96.7 & \blue{99.8} / \blue{99.9} 
&\red{99.9}	/ \red{100}\\
&Grid       & 77.2 / 90.5 & 61.4 / 78.0 & 85.9 / 94.3 & 89.9 / 96.2 & 56.2 / 77.4 & \red{99.5} / \red{99.8} & 45.1 / 68.1 & 97.1 / 99.1
&\blue{99.0} / \blue{99.7}\\
&Leather    & \red{100} / \red{100}   & \red{100} / \red{100}   & 64.1 / 90.0 & 98.2 / \blue{99.4} & \blue{99.9} / \red{100}  & 96.0 / 98.7 & 95.8 / 98.6 
& \red{100} / \red{100} & \red{100} / \red{100} \\
&Tile       & 98.0 / 99.2 & \blue{99.7} / \blue{99.9} & 50.0 / 85.9 & 98.5 / 99.5 & 98.4 / 99.5 & 98.7 / 99.5 & 91.9 / 97.3 & 99.3 / 99.8 
& \red{100} / \red{100}\\
&Wood       & \blue{99.5} / \blue{99.8} & 98.3 / 99.4 & 50.0 / 88.0 & \red{99.9} / \red{100}  & 97.5 / 99.1 & 97.6 / 99.2 & 98.2 / 99.4 & 98.5 / 99.6 
&98.8	/ 99.6\\
\cmidrule{2-11}
 \multirow{-17}{*}{\rotatebox{90}{\textbf{MVTec}\cite{mvtec}}} 
&\textbf{Mean} & 81.4 / 90.2 & 87.5 / 92.8 & 69.7 / 89.8 & 89.8 / 96.3 & 80.4 / 91.0 & 91.4 / 95.3 & 78.2 / 90.0 & \blue{96.5} / \blue{98.9} 
&\red{97.9}	/ \red{99.3}\\
\midrule\midrule
 
\multirow{4}*{\rotatebox{90}{\textbf{BTAD\cite{btad}}}} 
&01 & 95.1 / 98.0 & \red{99.8} / \red{99.9} & {98.7} / {99.5} & 98.2 / 99.2 & 96.0 / 98.6 & 93.1 / 97.5 & 96.4 / 98.3 & 92.2 / 97.9 &\blue{98.8}	/ \blue{99.6}\\
&02 & \blue{81.0} / \blue{96.7} & \red{87.9} / \red{98.1} & 50.0 / 93.5 & 70.2 / 95.8 & 70.8 / 95.0 & 61.4 / 90.9 & 75.2 / 96.2 & 78.8 / 96.4 &79.0 / 96.5\\
&03 & 99.4 / 94.3 & 99.4 / 94.2 & 57.5 / 55.4 & 99.3 / 94.0 & 95.6 / 69.6 & 99.5 / 96.7 & 99.3 / 90.6 & \blue{99.8} / \blue{98.0} &\red{100}	/ \red{99.5}\\
\cmidrule{2-11}
&\textbf{Mean} & 91.8 / 96.3 & \red{95.7} / 97.4 & 68.8 / 82.8 & 89.2 / 96.4 & 87.5 / 87.7 & 84.7 / 95.0 & 90.3 / 95.0 & {92.2} / \blue{97.9} &\blue{92.6} / \red{98.5}\\
\midrule\midrule
\multirow{13}*{\rotatebox{90}{
\textbf{VisA}\cite{visa}}}
&Candle      & 91.0 / 92.9 & 81.7 / 76.3 & 50.0 / 75.0 & 66.4 / 80.3 & 54.9 / 54.9 &  88.0 / 88.5 & 92.3 / 93.4 &\blue{96.6} / \blue{97.0}	&\red{96.8} / \red{97.1}\\
&Capsules    & 58.3 / 71.5 & 59.2 / 69.6 & 53.0 / 81.0 & \red{95.4} / \red{96.0} & 52.9 / 65.6 &  \blue{82.5} / \blue{90.7} & 76.2 / 85.3 &73.8 / 85.5	&79.0 / 89.0\\
&Cashew      & 91.8 / 95.7 & 83.0 / 91.2 & 53.0 / 84.3 & \red{96.0} / \red{98.0} & 82.2 / 89.6 &  65.4 / 81.0 & \blue{94.1} / \blue{97.0} &93.6 / 96.8	&93.7 / 96.7\\
&Chewinggum  & 98.9 / 99.5 & 87.2 / 94.2 & 53.5 / 84.5 & 98.3 / 99.3 & 90.2 / 95.5 &  94.0 / 97.5 & 97.1 / 98.8 &\blue{99.0} / \blue{99.5}	&\red{99.3} / \red{99.7}\\
&Fryum       & 88.0 / \blue{94.7} & 84.8 / 89.2 & 51.5 / 83.8 & \red{94.0} / \red{97.3} & 65.9 / 82.1 &  86.3 / 93.8 & 88.0 / 94.1 &\blue{88.4} / 94.5	&86.9 / 93.8\\
&Macaroni1   & 67.0 / 65.6 & 79.9 / 71.6 & 50.0 / 75.0 & 85.5 / 87.2 & 56.0 / 58.1 &  81.6 / 81.6 & 84.7 / 87.3 &\blue{89.3} / \blue{90.0}	&\red{91.9} / \red{91.6}\\
&Macaroni2   & 26.4 / 38.1 & 56.4 / 52.7 & 50.5 / 62.9 & 66.5 / 61.7 & 41.4 / 43.0 &  68.5 / 73.0 & 75.0 / 77.0 &\blue{82.1} / \blue{82.3}	&\red{84.1} / \red{86.5}\\
&Pcb1        & 91.2 / 90.1 & 77.0 / 72.6 & 50.0 / 75.0 & \blue{94.5} / \blue{94.7} & 88.9 / 86.6 &  69.6 / 71.2 & 93.4 / 94.4 &94.3 / 93.6	&\red{95.8} / \red{94.8}\\
&Pcb2        & 70.8 / 73.4 & 75.9 / 74.8 & 52.5 / 76.3 & \red{95.3} / \red{96.4} & 62.5 / 66.5 &  85.1 / 85.1 & 90.0 / 91.7 &91.9 / 93.0	&\blue{94.1} / \blue{94.6}\\
&Pcb3        & 54.0 / 61.8 & 69.5 / 63.3 & 53.0 / 76.4 & \red{95.3} / \red{95.5} & 72.4 / 73.6 &  84.9 / 85.8 & 91.3 / 93.4 &85.2 / 86.2	&\blue{91.9} / \blue{92.6}\\
&Pcb4        & 89.6 / 87.5 & 89.7 / 87.5 & 50.0 / 74.9 & \blue{99.4} / \blue{99.3} & 82.5 / 81.5 &  92.4 / 92.5 & 99.1 / 99.2 &99.3 / 99.2	&\red{99.5} / \red{99.5}\\ 
&Pipe Fryum  & 82.1 / 89.2 & 93.0 / 96.6 & 52.0 / 84.0 & \red{97.3} / \red{98.8} & 78.3 / 88.2 &  83.0 / 88.3 & 89.0 / 94.9 &\blue{96.4} / 98.2	&\red{97.3} / \blue{98.5}\\
\cmidrule{2-11}
&\textbf{Mean} &75.8 / 80.0 & 78.1 / 78.3 & 51.6 / 77.8 & 90.3 / 92.0 & 69.0 / 73.8 & 81.8 / 85.8 & 89.2 / 92.2 &\blue{90.8} / \blue{93.0}	&\red{92.5} / \red{94.5}\\
\bottomrule
\end{tabular}}
\vspace{-15pt}
\end{table*}

\end{document}